\documentclass[a4paper,fleqn]{cas-sc}
\usepackage[table]{xcolor} 
\usepackage{booktabs}      
\usepackage{tabularx}      
\usepackage{caption}       
\usepackage{makecell}      

\newcommand{\covfull}{\textcolor{black!85}{\rule[0.12ex]{1.0ex}{1.0ex}}}   
\newcommand{\covpart}{\textcolor{black!60}{$\circ$}}                        
\newcommand{\covnone}{\textcolor{black!35}{\,--\,}}                         

\definecolor{headerblue}{HTML}{E6F0FA}
\usepackage[numbers,sort&compress]{natbib}

\def\tsc#1{\csdef{#1}{\textsc{\lowercase{#1}}\xspace}}
\tsc{WGM}
\tsc{QE}
\tsc{EP}
\tsc{PMS}
\tsc{BEC}
\tsc{DE}

\begin{document}
\let\WriteBookmarks\relax
\def\floatpagepagefraction{1}
\def\textpagefraction{.001}

\shorttitle{Review of Sperm Analysis by Computer Vision and Deep Learning}

\shortauthors{Runwei Guan et~al.}

\title [mode = title]{Deep Learning for Semen Analysis in Male Infertility: Computer Vision, Multimodal Fusion, and Clinical Translation}                      



%
\author[3,4]{Runwei Guan}[
style=chinese,
]

\fnmark[1]
\ead{runwayrwguan@hkust-gz.edu.cn}
\credit{Conceptualization, Paper Writing and Correction}

\author[3]{Shaofeng Liang}[
style=chinese,
]
\fnmark[1]
\ead{shawnsfliang@hkust-gz.edu.cn}
\credit{Paper Writing, Review and Correction}

\author[5]{Jiacheng Weng}[
style=chinese,
]
\fnmark[1]
\ead{wj411411@163.com}
\credit{Conceptualization and Paper Writing}

\author[4]{Xiaoyi Gu}[
style=chinese,
]
\ead{xiaoyigu60@gmail.com}
\credit{Conceptualization}

\author[6]{Jia Weng}[
style=chinese,
]
\ead{jianice26211@163.com}
\credit{Paper Review and Correction}


\author[10]{Daizong Liu}[
style=chinese,
]
\ead{daizongliu@whu.edu.cn}
\credit{Paper Writing, Review and Correction}

\author[9]{Duo Pan}[
style=chinese,
]
\ead{dpan@srils.ac.cn}
\credit{Paper Review and Correction}

\author[8]{Qingxin Zhang}[
style=chinese,
]
\ead{qx.zhang@berkeley.edu}
\credit{Paper Review and Correction}

\author[2]{Xiao Liang}[
style=chinese,
]
\ead{heroshaw@hotmail.com}
\credit{Paper Review and Correction}


\author%
[7]
{Weiping Ding}[style=chinese]
\ead{ding.wp@ntu.edu.cn}
\credit{Paper Review and Correction}

\author[2]{Suoyu Zhu}[style=chinese]
\cormark[2]
\ead{zhusuoyu1104@163.com}
\credit{Paper Review, Correction and Supervision}

\author[1]{Ming Yuan}[style=chinese]
\cormark[1]
\ead{yuan__ming@163.com}
\credit{Paper Review and Correction}

\author[1]{Yanhua Fei}[style=chinese]
\cormark[3]
\ead{fertitechai@gmail.com}
\credit{Supervision}

\affiliation[1]{organization={Department of Gynaecology and Obstetrics, The Affiliated Jiangyin Hospital of Nantong University},
	city={Jiangyin},
	country={China}}

\affiliation[2]{organization={Department of Oncology, the Affiliated Jiangyin Hospital of Nantong University},
	city={Jiangyin},
	country={China}}

\affiliation[3]{organization={Thrust of AI, Hong Kong University of Science and Technology (Guangzhou)},
	city={Guangzhou},
	country={China}}

\affiliation[4]{organization={FertiTech AI},
	city={Shanghai},
	country={China}}

\affiliation[5]{organization={Department of Oncology, Suzhou Xiangcheng People's Hospital},
	city={Suzhou},
	country={China}}

\affiliation[6]{organization={Department of Biological Sciences and Bioinformatics, School of Science, Xi'an Jiaotong-Liverpool University},
	city={Suzhou},
	country={China}}

\affiliation[7]{organization={School of Artificial Intelligence and Computer Science, Nantong University},
	city={Nantong},
	country={China}}

    
\affiliation[8]{organization={Department of Electrical Engineering and Computer Sciences, University of California, Berkeley},
	city={Berkeley},
	country={USA}}

\affiliation[9]{organization={Sycamore Research Institute of Life Sciences},
	city={Shanghai},
	country={China}}

\affiliation[10]{organization={Institute for Math \& AI, Wuhan University},
	city={Wuhan},
	country={China}}


\cortext[cor1]{Corresponding author}

\fntext[fn1]{These authors contributed equally to this work and are co-first authors.}

\begin{abstract}
Male infertility contributes substantially to the global infertility burden, and sperm analysis remains central to diagnosis, treatment planning, and assisted reproductive technology. Conventional semen evaluation, however, is labor-intensive, operator-dependent, and limited by inter- and intra-observer variability, motivating the development of objective and reproducible computational approaches. This review provides a comprehensive and perspective-oriented synthesis of artificial intelligence-driven sperm analysis, with a focus on computer vision, deep learning, multimodal fusion, robustness, and clinical translation. We first review task-specific methods for sperm detection and counting, tracking-based motility assessment, semantic and instance segmentation, morphology and defect classification, functional assessment, and genetic integrity evaluation. We then summarize public datasets, benchmarks, evaluation metrics, and emerging multimodal strategies that integrate microscopic images, time-lapse videos, CASA-derived parameters, DNA integrity assays, and clinical metadata. Beyond algorithmic performance, we discuss key barriers to real-world deployment, including data scarcity, cross-center domain shift, annotation inconsistency, interpretability, uncertainty calibration, privacy-preserving learning, and workflow integration. Finally, we outline a staged clinical translation roadmap spanning technical standardization, multicenter retrospective validation, silent prospective evaluation, human-in-the-loop clinical testing, ART outcome validation, regulatory approval, and post-market monitoring. By organizing the field from task-specific visual recognition to trustworthy multimodal reproductive intelligence, this review highlights both the progress and the unresolved challenges required to translate AI-driven sperm analysis into clinically meaningful decision support.

\end{abstract}



\begin{keywords}
Sperm analysis \sep Computer vision \sep Deep learning \sep Multimodal fusion \sep Male infertility \sep Assisted reproductive technology \sep Trustworthy AI
\end{keywords}

\maketitle

\section{Introduction}

Male infertility remains a significant global health concern, contributing to approximately 40--50\% of all infertility cases among couples \cite{patel2018prediction}. Accurate and timely diagnosis is paramount for effective management and treatment, particularly in the context of assisted reproductive technologies (ART) such as in-vitro fertilization (IVF) and intracytoplasmic sperm injection (ICSI). The cornerstone of male fertility assessment has traditionally been manual semen analysis, a procedure that evaluates key parameters including sperm concentration, motility, and morphology according to guidelines set by organizations such as the World Health Organization (WHO). Throughout this review, unless otherwise stated, WHO parameters and morphological criteria refer to the current sixth edition of the WHO laboratory manual \cite{who2021laboratory,patel2018prediction}; because reference limits and morphological definitions have evolved across editions, we note the applicable standard whenever it materially affects the interpretation of a labelled dataset or reported result. However, conventional semen evaluation is not only operator-dependent but also fragmented across multiple sources of evidence, including microscopic morphology, dynamic motility patterns, CASA-derived kinematic parameters, DNA integrity assays, and patient-level clinical information. These heterogeneous signals are often interpreted separately in routine practice, limiting the ability to generate a unified, reproducible, and clinically actionable profile of sperm quality \cite{chang2024p}. Therefore, the central challenge is no longer merely to automate manual observation, but to integrate complementary visual, temporal, molecular, and clinical information into objective sperm assessment \cite{you2021machine}.

\begin{figure}
    \centering
    \includegraphics[width=0.998\linewidth]{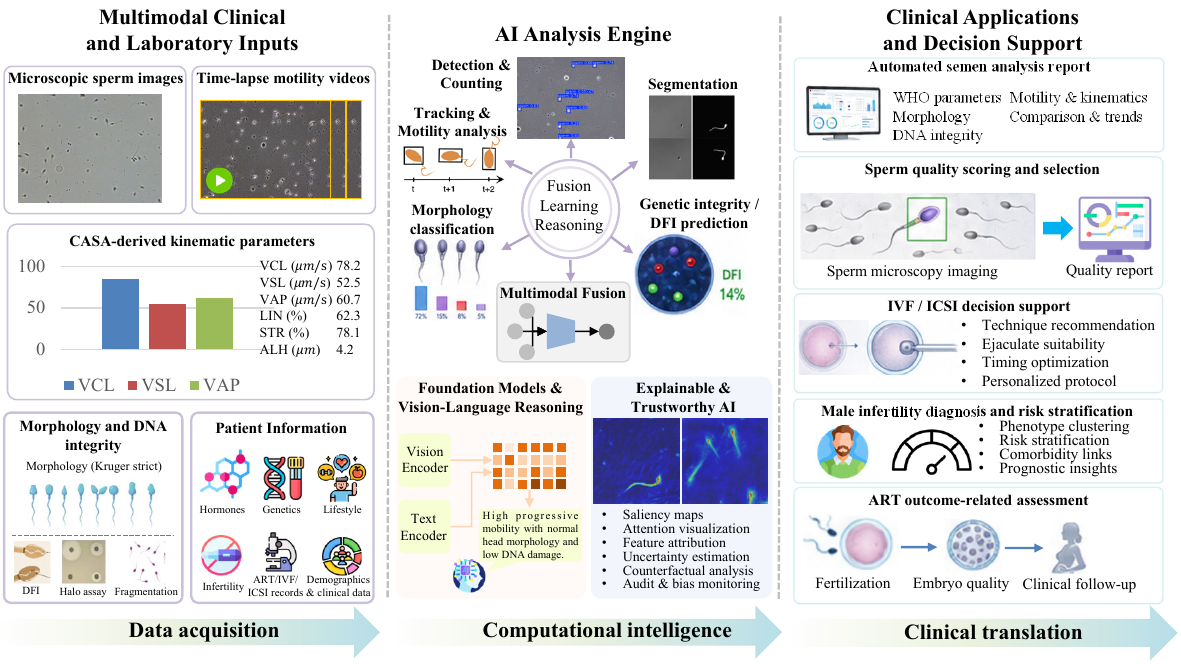}
    \vspace{-4mm}
    \caption{Translational framework of AI-driven sperm analysis. This figure summarizes how multimodal inputs, AI analysis modules, and clinical decision-support outputs are connected to enable automated, interpretable, and clinically translatable sperm assessment.}
    \label{fig:teaser}
\end{figure}

In recent years, the rapid advancements in computer vision (CV) and deep learning (DL) have revolutionized various fields, particularly biomedical imaging and diagnostics. These powerful computational paradigms offer unprecedented opportunities to overcome the limitations of traditional sperm analysis by automating complex tasks, enhancing precision, and reducing human bias. Deep learning models, with their ability to learn intricate patterns from vast datasets, are proving to be particularly adept at analyzing microscopic images and videos of sperm, enabling objective quantification of parameters that were previously difficult to assess consistently. The integration of CV and DL into sperm analysis promises to transform diagnostic workflows, improve patient outcomes, and facilitate more personalized reproductive medicine \cite{gongora2025new, abou2023artificial}.

Most existing AI studies in this field have been organized around isolated visual tasks, such as sperm detection, tracking, segmentation, or morphology classification. While these task-specific models have substantially improved the objectivity and efficiency of semen analysis, clinical fertility assessment requires a broader integration of complementary evidence. For example, motility trajectories provide dynamic functional information, morphology reflects structural normality, DNA fragmentation and other integrity assays capture molecular risk, and clinical metadata contextualize laboratory findings within patient-specific reproductive histories. As a result, AI-driven sperm analysis should be viewed as a multimodal information-fusion problem, in which heterogeneous data streams must be encoded, aligned, fused, interpreted, and validated across clinical contexts.

Several reviews have previously explored AI, machine learning, and CASA-related methods in reproductive medicine and sperm analysis. As Table \ref{tab:review_comparison} shows, these works have established important foundations for understanding task-specific automation and sperm selection; however, most of them remain organized around either individual computer-vision tasks, classical CASA pipelines, or general reproductive AI applications. Less attention has been paid to how heterogeneous laboratory and clinical signals can be fused, how such models generalize across devices and centers, and how AI outputs should be validated and integrated into clinical decision-making. For instance, \cite{abou2023artificial} is the first comprehensive overview of sperm deep learning models, systematically organized according to the CV task line, and summarizes 14 public datasets and clinical translation roadmaps. However, it offers relatively limited technical characterization of the individual model families and does not curate the datasets or model-evolution timeline into a single indexed reference point. \cite{schmeis2024predicting} is the first to quantitatively benchmark 43 ML or ANN models for male-infertility prediction. The limitations of it include high between-study heterogeneity, only seven ANN-specific papers, and the absence of a shared open-source benchmark that prevents direct cross-model comparison and clinical generalization. \cite{cherouveim2023artificial} provided a systematic review specifically on AI for sperm selection, highlighting its potential for objectivity and processing large datasets. Similarly, \cite{you2021machine} discussed machine learning for sperm selection, emphasizing its suitability for identifying promising gametes from vast populations. \cite{zhao2022survey} offered a survey of Computer-Assisted Sperm Analysis (CASA) methods since 1988, focusing on sperm detection and tracking, providing a historical perspective on automated techniques. \cite{dai2021advances} presented a broader overview of advances in sperm analysis, encompassing techniques beyond CV and DL, such as digital holography and super-resolution microscopy. \cite{louis2021review} is the first review dedicated to deep-learning-based computer vision in IVF embryo assessment, systematically analyzing 21 CNN studies spanning cell counting, detection, grading and outcome prediction, and highlighting the superiority of time-lapse over static morphology. Nevertheless, its limitations include a restricted English-language search, small sample size, absence of a unified benchmark or prospective clinical validation, and no discussion of cross-center domain adaptation or model interpretability.

\begin{table*}[t]
    \centering
    \setlength{\tabcolsep}{5.2pt}
    \renewcommand{\arraystretch}{1.18}
    \caption{Comparison with representative reviews on AI-based sperm analysis. The upper block reports \emph{verifiable scope coverage} (\covfull{}~fully covered, \covpart{}~partially/briefly covered, \covnone{}~not covered), judged from the presence of a corresponding dedicated treatment in each review. The lower block reports \emph{author-assessed depth} on a coarse three-level scale ($\star$ minimal, $\star\star$ partial, $\star\star\star$ substantial). Depth ratings are qualitative and inevitably involve some subjectivity; they are intended to summarize relative emphasis rather than to provide a precise metric. ``Curated dataset \& method index'' denotes a single traceable consolidation of dataset access points, model lineage, and per-study evaluation settings (compiled here in Sections~\ref{sec:datasets}--\ref{sec:metrics}); it does \emph{not} refer to a released re-implementation benchmark, which remains an open community need (Section~\ref{sec:emerging}).}
    \label{tab:review_comparison}
    \begin{tabular}{l|ccccccc|c}
    \toprule
      \multirow{2}{*}{\textbf{Aspect}}  &  \multicolumn{7}{c|}{\textbf{Prior reviews}} & \multirow{2}{*}{\textbf{Ours}} \\
         \cmidrule(lr){2-8}
         & \cite{abou2023artificial} & \cite{schmeis2024predicting} & \cite{you2021machine} & \cite{dai2021advances} & \cite{louis2021review} & \cite{cherouveim2023artificial} & \cite{zhao2022survey} & \\
    \midrule
    \multicolumn{9}{l}{\textit{Scope of coverage (verifiable)}}\\
    \midrule
    Task line: detection / tracking / segmentation / morphology & \covpart & \covnone & \covpart & \covpart & \covpart & \covpart & \covfull & \covfull \\
    Functional \& genetic-integrity assessment (DFI, ZP, AR) & \covpart & \covnone & \covpart & \covfull & \covnone & \covpart & \covpart & \covfull \\
    Public datasets \& benchmarks inventory & \covfull & \covfull & \covpart & \covnone & \covfull & \covpart & \covfull & \covfull \\
    Model-evolution timeline & \covnone & \covnone & \covnone & \covnone & \covnone & \covnone & \covnone & \covfull \\
    Multimodal-fusion framework (formalized) & \covpart & \covnone & \covpart & \covpart & \covnone & \covnone & \covnone & \covfull \\
    Robustness \& domain-shift analysis & \covnone & \covnone & \covpart & \covpart & \covnone & \covnone & \covpart & \covfull \\
    Clinical-translation roadmap (staged) & \covpart & \covpart & \covpart & \covpart & \covpart & \covpart & \covpart & \covfull \\
    Regulatory \& ethical considerations (IVDR / AI~Act / FDA) & \covnone & \covnone & \covnone & \covnone & \covnone & \covnone & \covnone & \covfull \\
    Evidence-tiering \& reporting caveats & \covnone & \covpart & \covnone & \covpart & \covnone & \covpart & \covnone & \covfull \\
    Curated dataset \& method index & \covnone & \covpart & \covnone & \covnone & \covpart & \covpart & \covpart & \covfull \\
    \midrule
    \multicolumn{9}{l}{\textit{Depth (author-assessed$^{\dagger}$)}}\\
    \midrule
    Technical depth for deep learning & $\star$ & $\star$ & $\star\star$ & $\star$ & $\star\star$ & $\star\star$ & $\star\star$ & $\star\star\star$\\
    Interdisciplinary (AI\,$\times$\,reproductive medicine) & $\star\star$ & $\star\star$ & $\star$ & $\star$ & $\star$ & $\star\star$ & $\star\star$ & $\star\star\star$\\
    Frontier / emerging-direction coverage & $\star$ & $\star$ & $\star$ & $\star$ & $\star$ & $-$ & $\star\star$ & $\star\star\star$\\
    \bottomrule
    \end{tabular}
    \vspace{0.4em}
    {\footnotesize $^{\dagger}$Depth ratings are the authors' qualitative judgment and may carry rating bias; they summarize relative emphasis within each review's stated scope and are not a quantitative benchmark.}
\end{table*}

Motivated by these gaps, as Figure \ref{fig:teaser} shows, this review positions AI-driven sperm analysis as an integrated multimodal reproductive-intelligence problem rather than a collection of independent computer-vision applications. Against this backdrop, our review makes the following contributions:

\begin{itemize}
    \item \textbf{A unified task-oriented synthesis.} We consolidate the full analytical pipeline: detection and counting, tracking-based motility assessment, semantic and instance segmentation, morphology and defect classification, and functional and genetic-integrity evaluation, within a single coherent framework, rather than treating these as isolated computer-vision problems.
    \item \textbf{A curated data and evaluation index.} We compile public datasets, annotation protocols, benchmarks, and task-specific evaluation metrics into a single traceable reference, together with a model-evolution timeline that situates each method family in its methodological lineage.
    \item \textbf{A formalized multimodal-fusion perspective.} We cast sperm assessment as an explicit information-fusion problem over heterogeneous modalities (microscopic images, time-lapse videos, CASA-derived kinematics, DNA-integrity assays, and clinical metadata), and organize early, intermediate, and late fusion strategies by their handling of missing modalities and uncertainty.
    \item \textbf{A robustness- and translation-centred critical analysis.} We critically examine domain generalization, interpretability, privacy-preserving learning, regulatory and ethical considerations, and staged clinical validation as prerequisites for trustworthy ART decision support, dimensions largely absent from prior reviews.
\end{itemize}

The remainder of this review is organized as follows. Section~\ref{sec:search} describes the search strategy and review scope. Sections~\ref{sec:detection_tracking}--\ref{sec:genetic} then present the task-oriented synthesis, covering sperm detection and tracking, segmentation, morphology and functional classification, and genetic-integrity assessment, respectively. Section~\ref{sec:datasets} surveys public datasets and benchmarks, and Section~\ref{sec:metrics} details the task-specific evaluation metrics. Section~\ref{sec:emerging} discusses multimodal fusion and other emerging paradigms, followed by scenario-based guidance for method selection in Section~\ref{sec:guidelines}. Section~\ref{sec:clinical} outlines the clinical-translation pathway, and Section~\ref{sec:robustness} provides a forward-looking analysis of model robustness and domain adaptation, before the concluding remarks in Section~\ref{sec:conclusion}.

\section{Search Strategy and Review Scope}
\label{sec:search}

\begin{figure}
    \centering
    \includegraphics[width=0.998\linewidth]{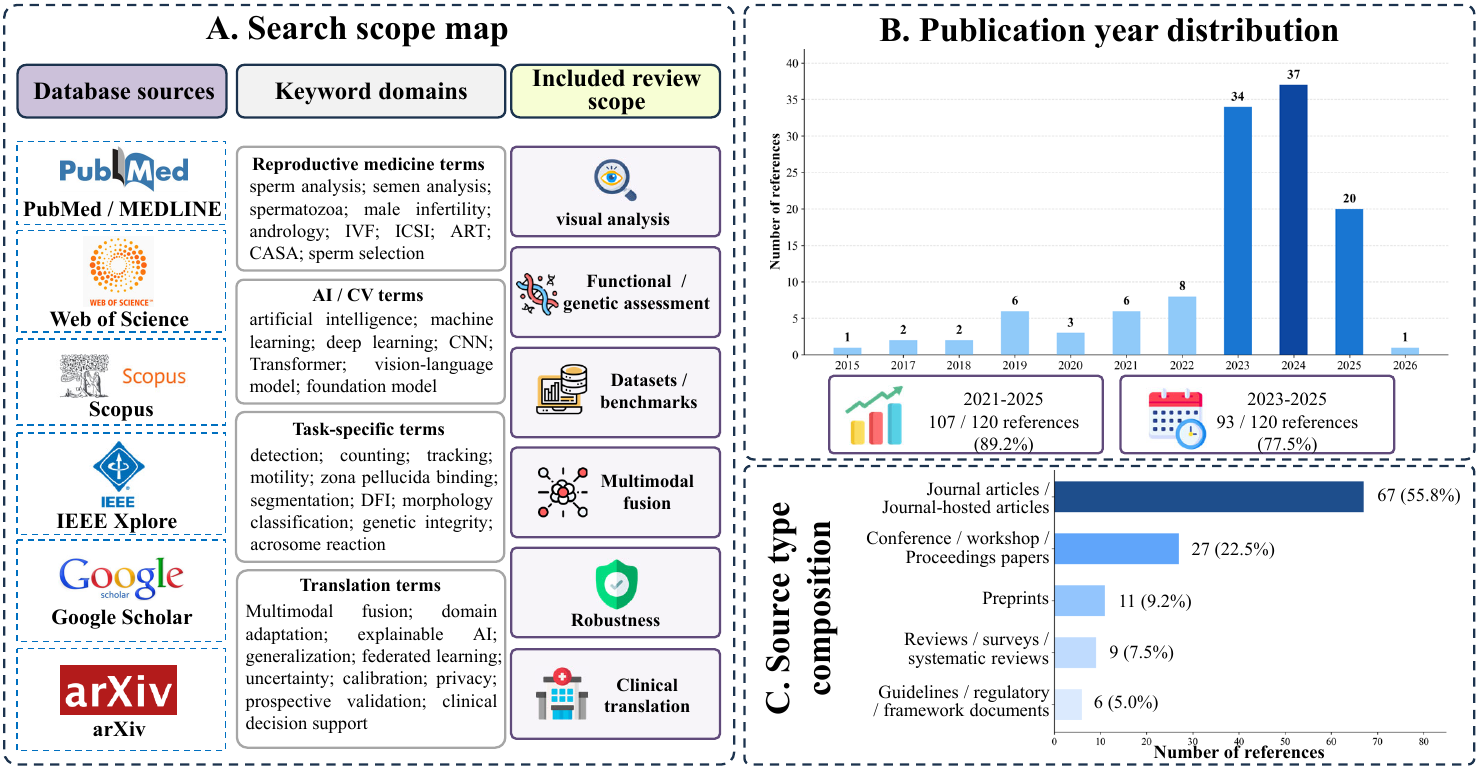}
    \vspace{-4mm}
    \caption{Search scope and reference landscape of AI-driven sperm analysis. The figure summarizes the semi-structured search scope, keyword domains, included thematic coverage, publication-year distribution, and source-type composition of the references reviewed in this article.}
    \label{fig:paper_stat}
\end{figure}

This review was designed as a narrative and perspective review rather than a formal systematic review or meta-analysis. The aim was to provide a task-oriented and translation-oriented synthesis of artificial intelligence methods for sperm analysis, with particular emphasis on computer vision, deep learning, multimodal fusion, robustness, and clinical deployment. A semi-structured literature search was conducted using major biomedical and engineering databases, including PubMed/MEDLINE, Web of Science, Scopus, IEEE Xplore, and Google Scholar. To capture rapidly evolving AI methodologies, relevant preprint servers and open-access repositories, including arXiv, medRxiv, and bioRxiv, were also screened when appropriate.


Studies were considered eligible if they met at least one of the following criteria: (i) they proposed, evaluated, or reviewed AI-based methods for sperm or semen analysis; (ii) they introduced datasets, benchmarks, or annotation protocols relevant to sperm detection, tracking, segmentation, morphology classification, motility analysis, DNA integrity assessment, or sperm selection; (iii) they investigated functional, genetic, or molecular sperm-quality indicators that could be analyzed or supported by AI; or (iv) they provided methodological foundations relevant to multimodal fusion, model robustness, interpretability, privacy-preserving learning, or clinical translation. Both journal articles and conference papers were included when they provided clear methodological or translational relevance. Preprints were considered selectively when they represented emerging AI paradigms or rapidly developing areas not yet fully covered by peer-reviewed literature.

We excluded studies that were not directly related to sperm, semen, male infertility, reproductive medicine, or transferable AI methodology; papers without sufficient methodological detail; duplicate reports; non-research commentaries without technical or clinical relevance; and studies in which AI was only mentioned superficially without being central to the analysis. Given the heterogeneity of tasks, datasets, endpoints, imaging modalities, and evaluation protocols, no quantitative meta-analysis was performed. Instead, the literature was organized according to analytical tasks, data resources, model families, evaluation metrics, multimodal fusion strategies, and clinical translation challenges. This scope allows the review to identify not only algorithmic progress but also unresolved barriers to reproducible, generalizable, and clinically trustworthy AI-driven sperm analysis.



\begin{figure}
    \centering
    \includegraphics[width=0.49\linewidth]{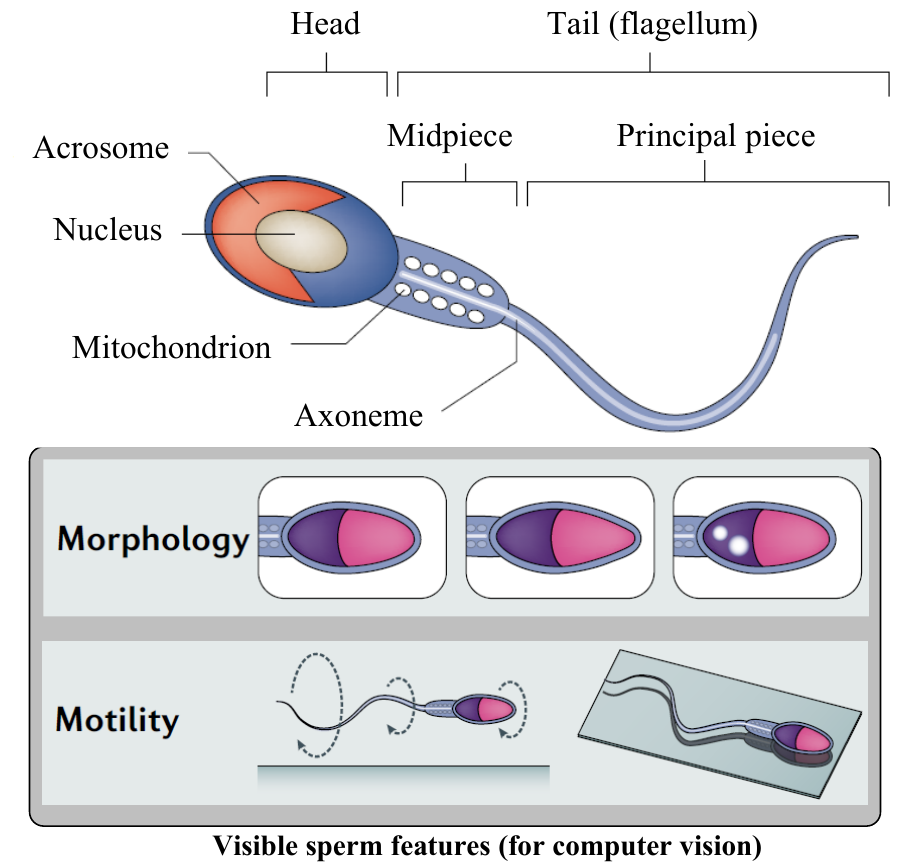}
    \vspace{-2mm}
    \caption{Detailed structure of sperm and visible sperm features discernible by computer vision (part of the figure is adapted from You et al. \cite{you2021machine}).}
    \label{fig:sperm_review}
\end{figure}

\begin{figure}
    \centering
    \includegraphics[width=0.72\linewidth]{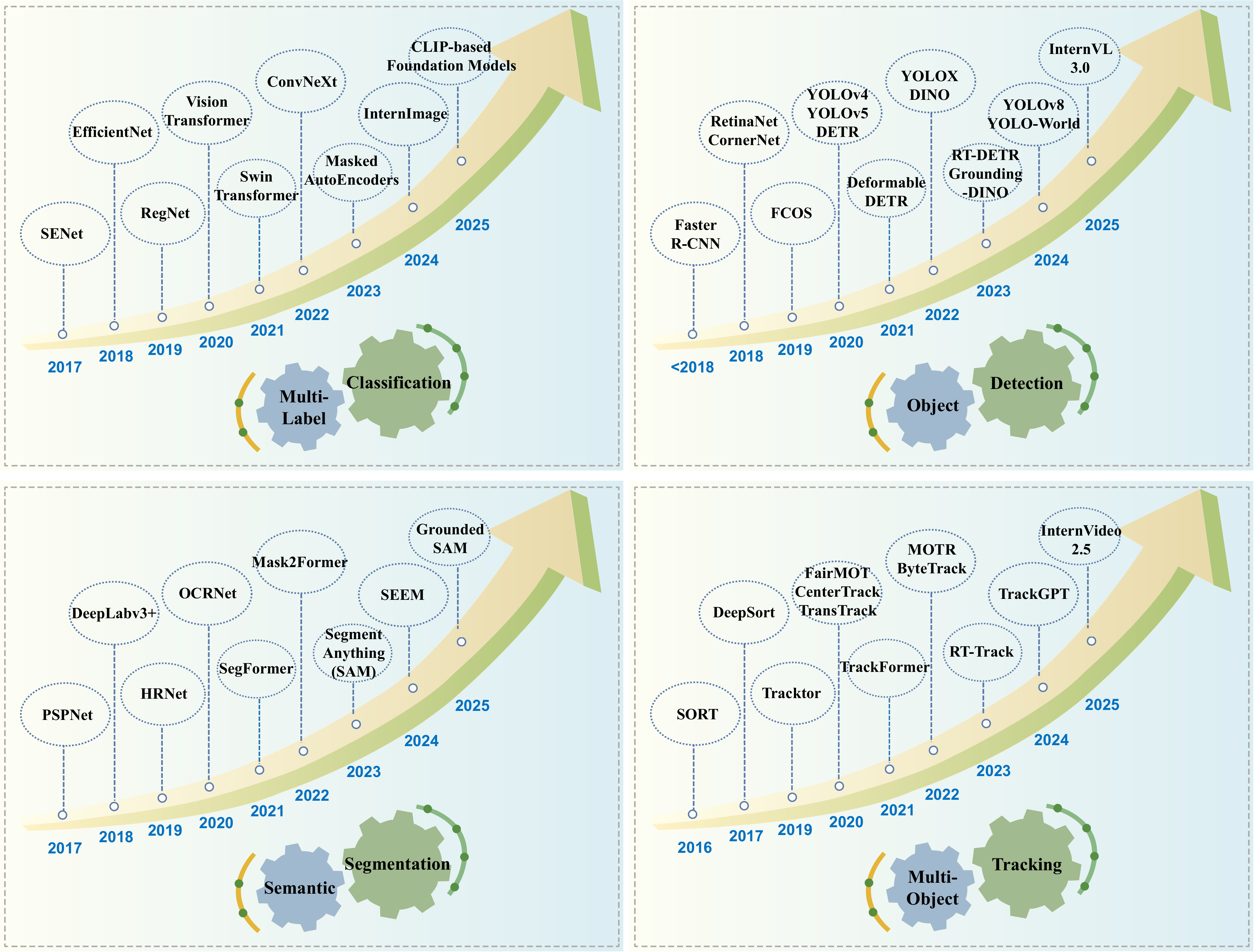}
    \vspace{-2mm}
    \caption{Evolution of representative computer-vision architectures relevant to AI-driven sperm analysis. The timeline summarizes major model families for classification, detection, segmentation, and multi-object tracking, which provide methodological foundations for sperm morphology analysis, localization, motility tracking, and related visual tasks.}
    \label{fig:timeline}
\end{figure}

%

\section{Automated Sperm Detection and Tracking}
\label{sec:detection_tracking}

Accurate localization and temporal association of sperm cells (Figure \ref{fig:sperm_review}) constitute the foundational layer of computer-assisted semen analysis. Object detection provides spatial coordinates and bounding box representations of individual spermatozoa within a given frame, while multi-object tracking (MOT) extends this capability by establishing cross-frame identity correspondences, thereby enabling the reconstruction of motility trajectories. These two tasks are tightly coupled: reliable detection underpins robust tracking, and conversely, temporal consistency afforded by tracking can refine detection performance in challenging scenarios such as motion blur, overlapping cells, and variable illumination. In this section, we review the deep learning approaches developed for sperm detection and tracking, followed by a discussion of end-to-end motility classification methods that bypass explicit trajectory extraction in favor of direct predictive modeling.

\begin{figure}
    \centering
    \includegraphics[width=0.998\linewidth]{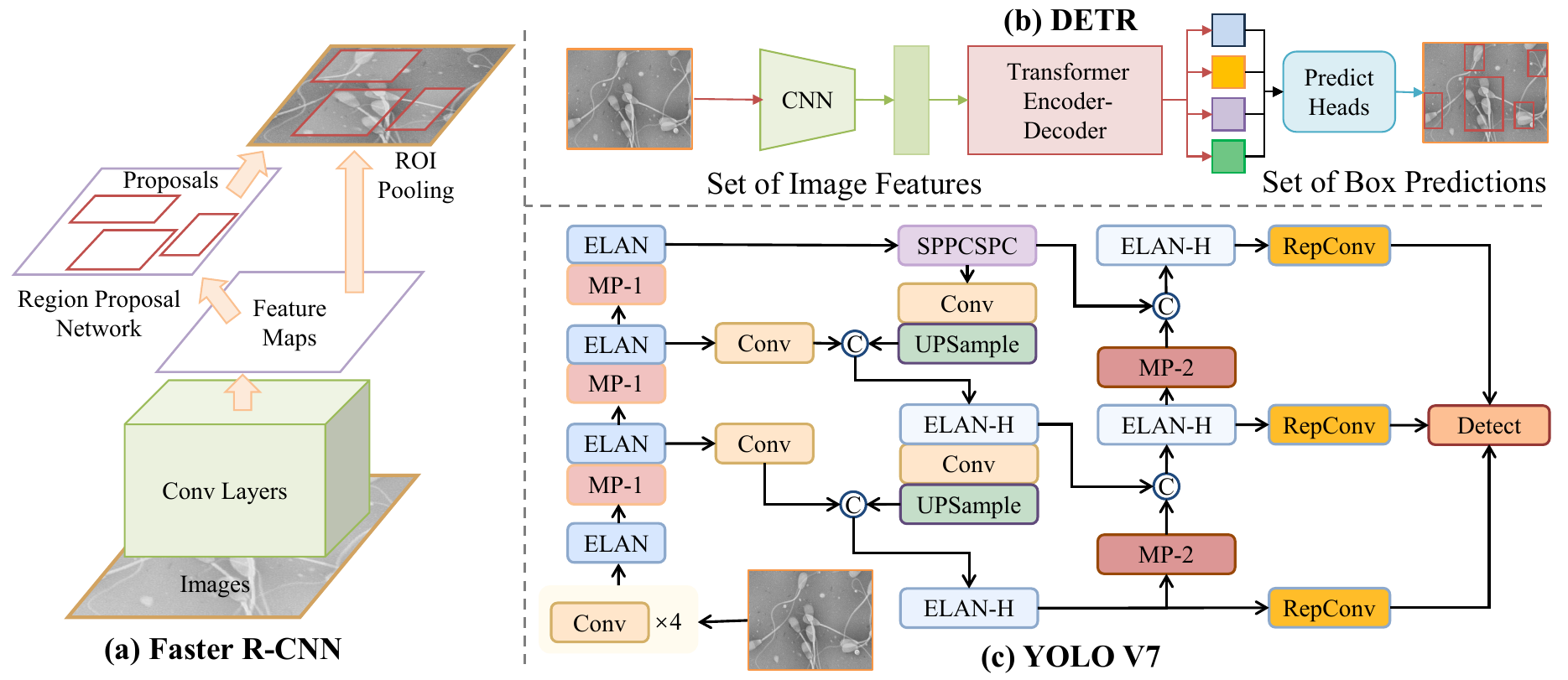}
    \vspace{-4mm}
    \caption{Overview of sperm object detection models with various paradigms.}
    \label{fig:detection}
\end{figure}

\subsection{Object Detection Models for Sperm}
\label{subsec:object_detection}

Object detection models have become the predominant approach for localizing sperm cells in microscopic video frames, offering a balance between computational efficiency and localization accuracy. As illustrated in Figure~\ref{fig:detection}, modern sperm detection pipelines leverage both single-stage and two-stage convolutional neural network architectures adapted to the unique challenges of microscopic imaging, including small object size, high cell density, and significant inter-frame motion.

A notable early contribution specifically targeting the small-object challenge in sperm imaging is TOD-CNN~\cite{zou2022tod}, a convolutional neural network architecture explicitly designed for tiny object detection in sperm video sequences. TOD-CNN achieved a detection performance of 85.60\% AP$_{50}$, demonstrating the feasibility of specialized architectures for densely packed microscopic scenes. To avoid conflating training scale with dataset scale, we note that the frequently quoted figure of ``$>$278{,}000 annotated objects'' is the \emph{aggregate} annotation count across the three subsets of the SVIA dataset (Table~\ref{tab:datasets}), not the annotation budget of any single detector; the exact per-method training scale should be read from each original paper. The data-intensive nature of such supervised detection approaches highlights the reliance of this domain on large-scale annotated microscopic datasets.

\begin{table}
    \centering
    \setlength\tabcolsep{2.0pt}
    \caption{Representative sperm detection methods and their characteristics}
    \scriptsize
    \begin{tabular}{c|cccc}
    \rowcolor{headerblue}
    \toprule
       \textbf{Methods}  & \textbf{Core Idea} & \textbf{Pros} & \textbf{Cons} & \textbf{Representative} \\
       \midrule
        YOLO  & Single-stage  & High speed, good accuracy, & Struggle with tiny objects & \multirow{2}{*}{\cite{lopuran2025hybrid,dobrovolny2023study,dobrovolny2022sperm}}\\\
        variants & real-time detection &  continuously improving &  or dense occlusions & \\
        \midrule
        \multirow{2}[2]{*}{Faster R-CNN} & Two-stage detection & Robust for diverse & Slower than single-stage & \multirow{2}{*}{\cite{yuzkat2023detection}}\\
         & (region proposal +classification) & object scales & high computation cost & \\
        \midrule
        Specialized CNNs & Custom architectures for & Optimized for tiny detection, & Requires specialized design, & \multirow{2}{*}{\cite{zou2022tod,chen2024active}} \\
        (e.g., TOD-CNN) & specific challenges & robust against impurities & not generalized & \\
        \midrule
        \multirow{2}[2]{*}{DETRs}  & query-based end-to-end & universal sequence prediction, & high computation cost, & \multirow{2}{*}{N/A}\\
         & detection without NMS & nice scalability & fixed object queries \\ 

    \bottomrule
    \end{tabular}
    \label{tab:detection_model}
\end{table}

The YOLO (You Only Look Once) family of single-stage detectors has gained considerable traction in sperm detection research due to its favorable speed-accuracy trade-off. A comparative study~\cite{yuzkat2023detection} systematically evaluated single-stage against two-stage detectors for sperm detection and concluded that YOLOv5 delivered the best overall performance among the tested configurations. Building upon this foundation, YOLOv5s-SA~\cite{zhu2023yolov5s} introduced architectural enhancements specifically tailored for sperm detection, incorporating the Shuffle Attention (SA) mechanism, depthwise separable convolutions, and multi-scale feature fusion. These modifications yielded a substantial improvement of 18.1\% over the baseline YOLOv3 architecture, underscoring the benefits of attention mechanisms and lightweight convolutions for resource-constrained clinical deployment. Furthermore, YOLOv5 has been successfully applied to sperm count estimation~\cite{dobrovolny2023study,dobrovolny2022sperm}, achieving 72.15\% mAP on the publicly available Visem dataset and demonstrating the versatility of detection frameworks beyond mere localization.

\begin{figure}
    \centering
    \includegraphics[width=0.92\linewidth]{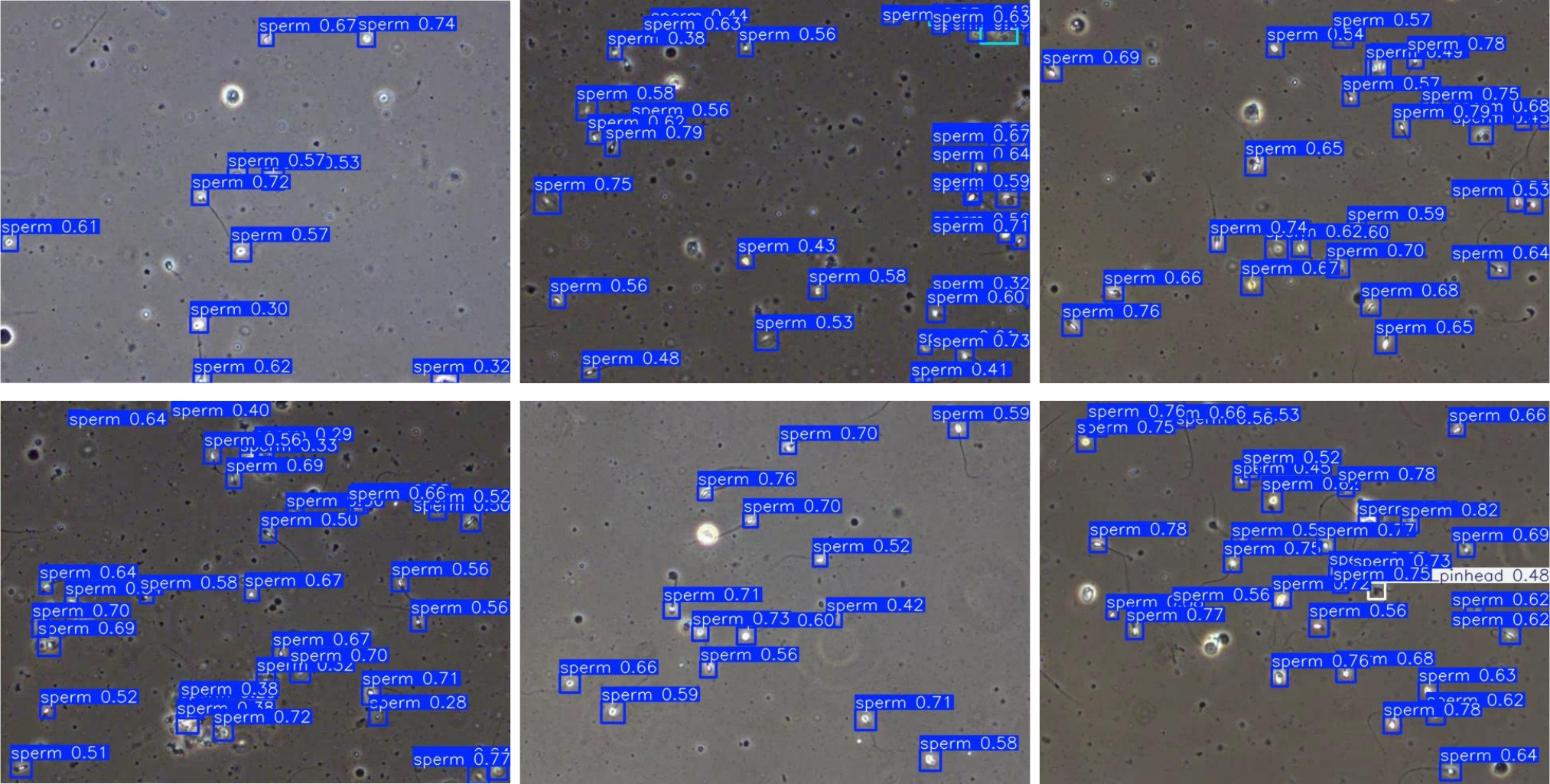}
    \vspace{-2mm}
    \caption{Samples of object detection for sperm.}
    \label{fig:detection_samples}
\end{figure}

Collectively, these developments reflect a broader trend toward adapting general-purpose detection architectures to the specific visual characteristics of spermatozoa. Table~\ref{tab:detection_model} provides a comparative summary of the detection methods discussed, highlighting their architectural choices. Representative detection results from selected methods are visualized in Figure~\ref{fig:detection_samples}, illustrating the quality of bounding box localization under varying imaging conditions.

\subsection{Multi-Object Tracking for Motility Analysis}
\label{subsec:tracking}

\begin{figure}
    \centering
    \includegraphics[width=0.998\linewidth]{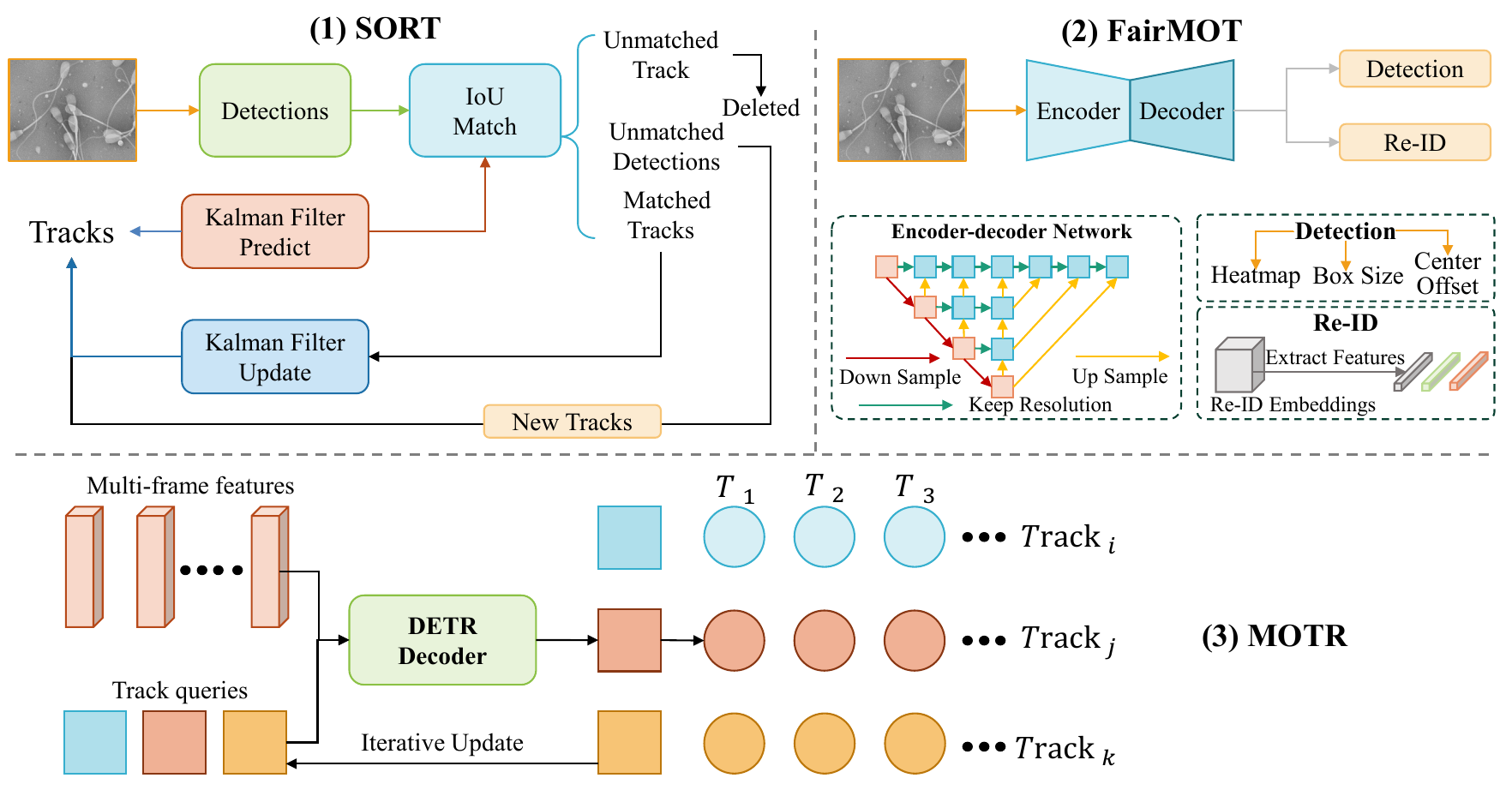}
    \vspace{-4mm}
    \caption{Overview of algorithm/model architectures for multi-object sperm tracking.}
    \label{fig:tracking}
\end{figure}

While detection localizes sperm within individual frames, multi-object tracking establishes temporal correspondences across consecutive frames, enabling the extraction of kinematic parameters essential for motility grading. The tracking task in semen analysis is particularly challenging due to the high density of sperm populations, frequent occlusions, similar visual appearances among cells, and rapid, erratic movement patterns. As shown in Figure~\ref{fig:tracking}, contemporary sperm tracking systems typically adopt a detection-followed-by-tracking paradigm, wherein deep learning-based detectors provide frame-wise detections that are subsequently associated through dedicated tracking algorithms.

One of the pioneering integrated approaches combined Faster R-CNN with the Enhanced Sperm Analysis (ESA) framework and the Temporal History Motion Analysis (THMA) module~\cite{somasundaram2021faster}. This system achieved a classification accuracy of 97.37\% and was capable of detecting sperm motion within 1.12 seconds, demonstrating the potential of two-stage detectors paired with temporal analysis for real-time clinical applications. The introduction of the VISEM-Tracking dataset~\cite{thambawita2023visem}, comprising 20 videos with manually annotated bounding boxes, has provided a valuable benchmark for evaluating tracking algorithms in this domain.

More recently, SpermYOLOv8E-TrackEVD~\cite{zhang2024sperm} proposed an enhanced YOLOv8 detector coupled with the SpermTrack-EVD association algorithm, achieving a Higher Order Tracking Accuracy (HOTA) of 74.303\% and Multiple Object Tracking Accuracy (MOTA) of 71.167\%. These metrics represent significant progress in balancing detection and association quality for densely populated microscopic scenes. Complementing this work, a hybrid architecture~\cite{lopuran2025hybrid} integrated UNet for segmentation, YOLOv8 for detection, and DeepSORT for tracking into a unified pipeline, reporting 98.21\% segmentation accuracy, 98.43\% detection mAP, and 90.41\% MOTA. This multi-stage fusion strategy illustrates how complementary deep learning modules can be orchestrated to address different aspects of the tracking problem simultaneously.

The ACTIVE framework~\cite{chen2024active,chen2023active} introduced a Dedicated Backbone Feature Extraction Network (DBFEN) combined with a Cross-Connected Feature Pyramid Network (CCFPN), achieving an AP$_{50}$ of 91.13\% for sperm detection and 59.64\% for debris detection. The ability to distinguish sperm from cellular debris and other artifacts is critical for accurate motility assessment, as impurities can confound both detection and subsequent kinematic analysis. In a similar vein, a complete Computer-Assisted Semen Analysis (CASA) system~\cite{tiab2024deep} integrated MobileNet for feature extraction, YOLOv5s for detection, and DeepSort for tracking, attaining 99.2\% classification accuracy, 96--99.5\% mAP@50, and 99\% Multiple Object Tracking Precision (MOTP). These results highlight the maturity of tracking-based approaches for clinical-grade semen analysis.

Reported tracking scores, however, must be read together with the acquisition physics that bound their clinical validity. Because spermatozoa swim rapidly and along curved paths, the CASA-derived kinematic parameters that motility grading depends on, curvilinear velocity (VCL), straight-line velocity (VSL), average path velocity (VAP), amplitude of lateral head displacement (ALH), and beat-cross frequency (BCF), are jointly determined by the video frame rate, the spatial calibration ($\mu$m per pixel at the given magnification), and stage temperature control. An insufficient frame rate undersamples the trajectory in the Nyquist sense: it systematically straightens curvilinear paths, deflates VCL and ALH, and inflates linearity, so that two systems reporting similar HOTA~\cite{luiten2021hota} or MOTA can still disagree on the very kinematics used for WHO grading. Similarly, uncalibrated magnification biases all velocity estimates by a constant factor, and imaging away from 37$^\circ$C alters the motility being measured rather than merely how it is observed. Consequently, a high identity-preservation score (HOTA/IDF1) is necessary but not sufficient for trustworthy motility analysis; frame rate, calibration, and temperature should be reported alongside tracking metrics, and cross-study kinematic comparisons are only meaningful when these acquisition settings are matched.

The evolution from conventional CASA systems~\cite{10.21661/r-559395,tang2023analysis,jorge2024specific} to deep learning-enhanced CASA architectures represents a paradigm shift in how motility data are acquired and processed. Traditional CASA systems rely on frame-differencing and threshold-based segmentation, which are sensitive to illumination changes and cannot reliably handle overlapping cells. In contrast, modern deep learning-based CASA platforms leverage learned representations that are robust to such variations. Hardware optimization studies~\cite{hernandez2023recasa} have further explored the deployment of these algorithms on resource-constrained devices, while comparative investigations using GoldCyto versus glass slide preparations~\cite{akal2023evaluation} have examined the influence of sample handling protocols on detection and tracking performance.

\subsection{End-to-End Motility Classification}
\label{subsec:e2e_motility}

\begin{figure}
    \centering
    \includegraphics[width=0.998\linewidth]{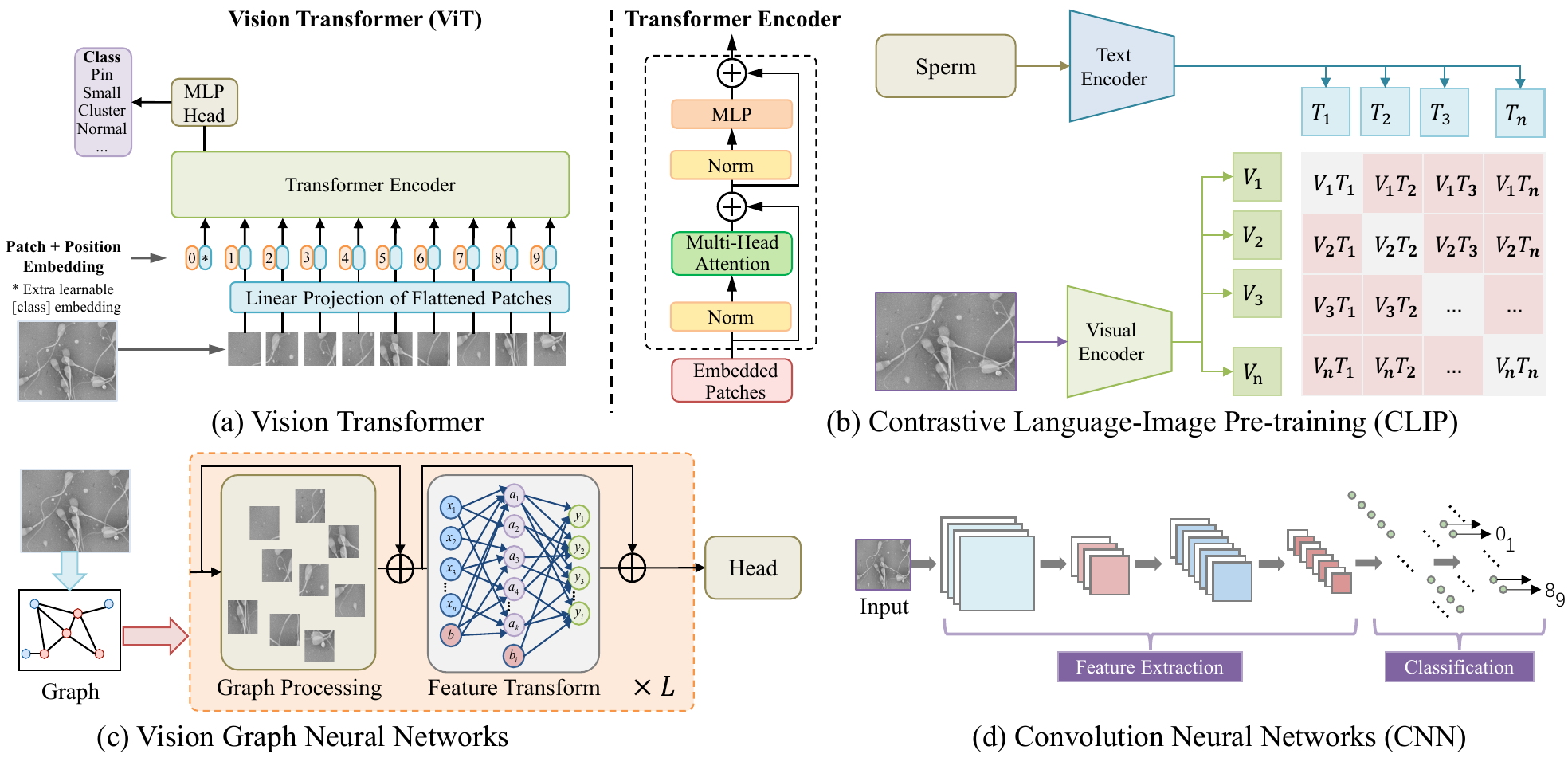}
    \vspace{-4mm}
    \caption{The architecture of the classification models that can be used for sperm morphology classification.}
    \label{fig:classification}
\end{figure}

An alternative to the explicit detect-track-classify pipeline is the end-to-end motility classification paradigm, wherein deep learning models directly predict World Health Organization (WHO) motility categories from raw or minimally processed video sequences without extracting intermediate trajectories. This approach streamlines the analysis workflow and avoids the propagation of errors from detection and tracking stages to the final motility assessment.

In one of the earliest explorations of this direction, a hybrid machine learning approach~\cite{hicks2019machine} combined linear regression with convolutional neural networks to analyze sperm motion videos. The authors found that deep learning-based predictions were not only faster than traditional tracking-based methods but also exhibited greater consistency across repeated evaluations, suggesting that end-to-end models can capture motion-relevant features that may be overlooked by explicit kinematic parameter extraction.

Subsequently, a deep convolutional neural network based on the ResNet-50 architecture~\cite{haugen2023sperm} was developed to directly predict WHO motility categories from sperm video recordings. Evaluated through ten-fold cross-validation on 65 wet-mount semen videos, this approach achieved a mean absolute error (MAE) of 0.05 for three-class and 0.07 for four-class motility grading, with a Pearson correlation coefficient of $r = 0.88$ relative to expert assessments. The strong correlation between model predictions and human expert judgments indicates that end-to-end architectures can effectively learn discriminative spatiotemporal representations for motility characterization. These findings suggest that end-to-end classification may serve as a viable and efficient alternative to traditional tracking-based motility analysis, particularly in high-throughput screening settings where computational latency and annotation burden must be minimized.

\section{Deep Learning for Sperm Segmentation}
\label{sec:segmentation}

While object detection provides bounding box-level localization, semantic and instance segmentation offer pixel-level delineation of sperm structure, enabling precise morphometric analysis that is essential for assessing normalcy according to WHO criteria. Segmentation facilitates the extraction of biologically meaningful measurements such as head dimensions, acrosome coverage, midpiece length, and tail elongation---features that are critical for diagnosing teratozoospermia and other morphological abnormalities. The field has evolved from early thresholding and active contour methods to sophisticated deep learning architectures capable of both semantic segmentation (classifying each pixel as sperm or background) and instance segmentation (separating individual sperm cells and their component parts). As depicted in Figure~\ref{fig:segmentation}, the progression of segmentation methodologies has closely followed advances in general computer vision, with domain-specific adaptations addressing the unique challenges of sperm cell morphology. This section reviews semantic segmentation approaches, instance and part-level segmentation methods, and the emerging role of foundation models in sperm image analysis.

\subsection{Semantic Segmentation of Sperm}
\label{subsec:semantic_segmentation}

\begin{figure}
    \centering
    \includegraphics[width=0.998\linewidth]{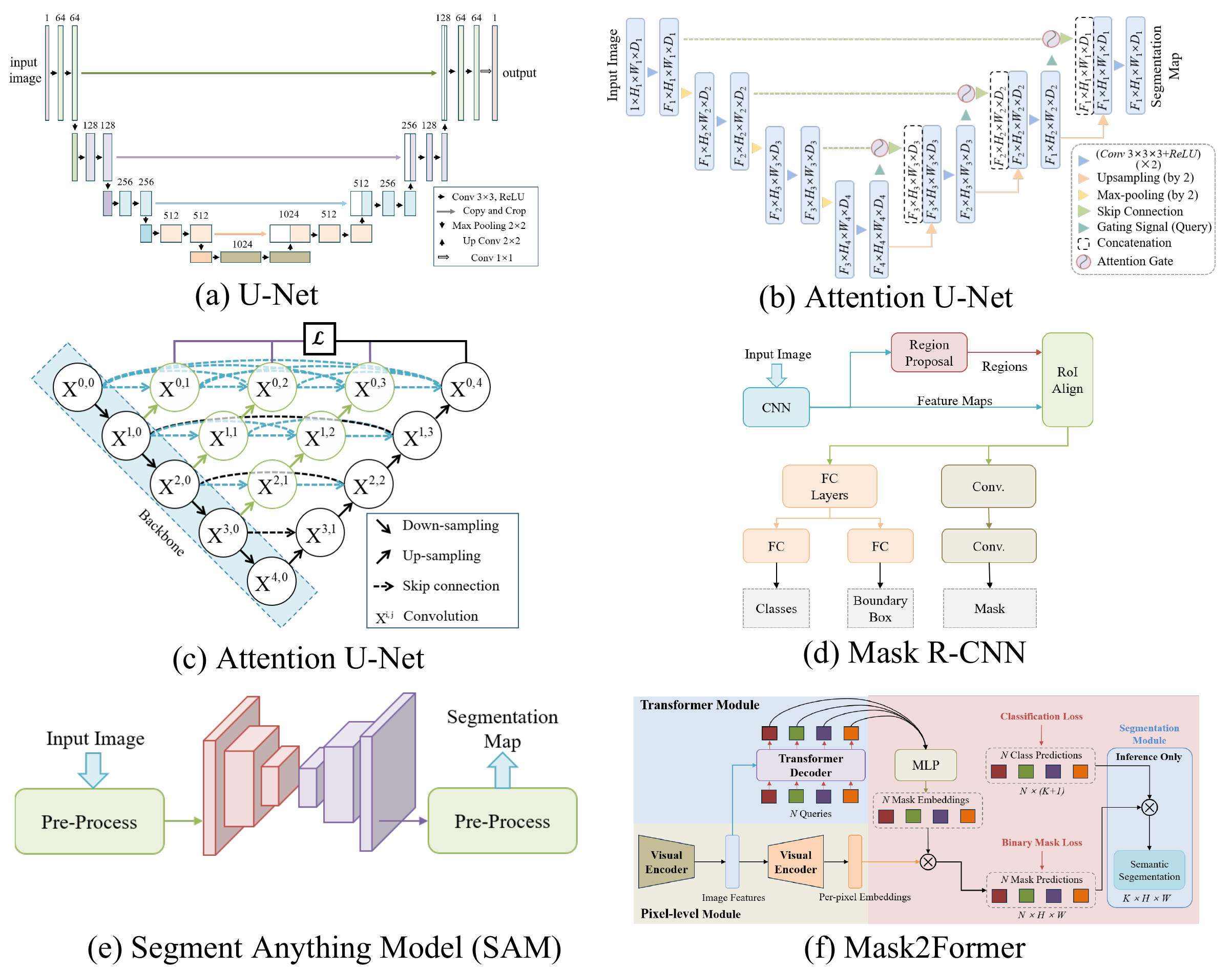}
    \vspace{-4mm}
    \caption{Overview of representative segmentation architectures applicable to sperm image analysis, including encoder--decoder networks (U-Net, Attention U-Net, UNet++), instance segmentation (Mask R-CNN), and foundation/transformer-based models (SAM, Mask2Former).}
    \label{fig:segmentation}
\end{figure}

Semantic segmentation approaches classify every pixel in a microscopic image as belonging to either sperm or background regions, providing a dense spatial mask that supports subsequent morphological quantification. The U-Net architecture and its variants have become the cornerstone of sperm semantic segmentation due to their encoder-decoder structure with skip connections, which effectively preserves fine-grained spatial details necessary for delineating slender sperm tails against noisy backgrounds.

An early application of U-Net to sperm cell segmentation~\cite{melendez2021sperm} demonstrated the architecture's suitability for this task, achieving 93\% accuracy, 88\% Intersection over Union (IoU), and 94\% Dice coefficient on sperm microscopy images. These results established U-Net as a strong baseline for pixel-level sperm analysis. Recognizing the limitations of the standard U-Net in capturing multi-scale features, subsequent work~\cite{lv2022improved} introduced an improved U-Net incorporating atrous (dilated) convolutions and custom feature enhancement modules. Evaluated on a dataset of 1,207 sperm images, this enhanced architecture achieved a Dice coefficient of 95.14\%, representing a notable improvement over the original formulation and demonstrating the value of expanded receptive fields for sperm morphology analysis.

\begin{table}
    \centering
    \setlength\tabcolsep{2.0pt}
    \caption{Representative sperm segmentation methods and their characteristics}
    \scriptsize
    \begin{tabular}{c|cccc}
    \rowcolor{headerblue}
    \toprule
       \textbf{Methods}  & \textbf{Core Idea} & \textbf{Pros} & \textbf{Cons} & \textbf{Representative} \\
       \midrule
        U-Net  & Encoder-decoder  & Capture fine details, & Struggle with overlapped objects, & \multirow{2}{*}{\cite{zhang2026dynamic,lv2022improved,hernandez2023deep}}\\
        variants & architecture &  relatively efficient &  depend on data quality & \\
        \midrule
        \multirow{2}[2]{*}{Mask R-CNN} & Combine object detection & Provide both bounding boxes & Slower than U-Net & \multirow{2}{*}{\cite{marin2021impact}}\\
         & with instance segmentation. & and pixel-level mask & high computation cost & \\
        \midrule
        Segment Anything & Foundation model for & Highly-generalizable, & Requires finetuning for , & \multirow{2}{*}{\cite{shi2024cs3}}\\
        Model (SAM) & zero-shot segmentation & with minimal training data & specific medical images & \\
    \bottomrule
    \end{tabular}
    \label{tab:segmentation_methods}
\end{table}

Beyond architectural modifications, researchers have also explored multi-task and transfer learning strategies to enhance segmentation quality. Hern\'andez-Herrera et al.~\cite{hernandez2023deep} employed U-Net to simultaneously segment sperm heads and flagella, achieving a Dice score of 0.81, and further integrated the segmentation masks with a ResNet50-based framework for interpretable bull fertility classification, illustrating how segmentation serves as an enabling module for downstream diagnostic tasks. This study was conducted on bovine spermatozoa; as with the other veterinary examples cited in this review, the morphological criteria and fertility endpoints differ from human semen analysis, so the segmentation methodology transfers more readily than the species-specific decision thresholds. In a complementary direction, transfer learning was leveraged to augment both U-Net and Mask R-CNN architectures for sperm segmentation~\cite{marin2021impact}, resulting in 95\% overlap rates and improved Dice coefficients compared to training from scratch. This finding highlights the utility of pre-trained representations, even when sourced from natural image datasets, for microscopic sperm imagery. Figure~\ref{fig:segmentation} illustrates the architectural diversity of models applied to sperm semantic segmentation tasks.

\begin{figure*}
    \centering
    \includegraphics[width=0.99\linewidth]{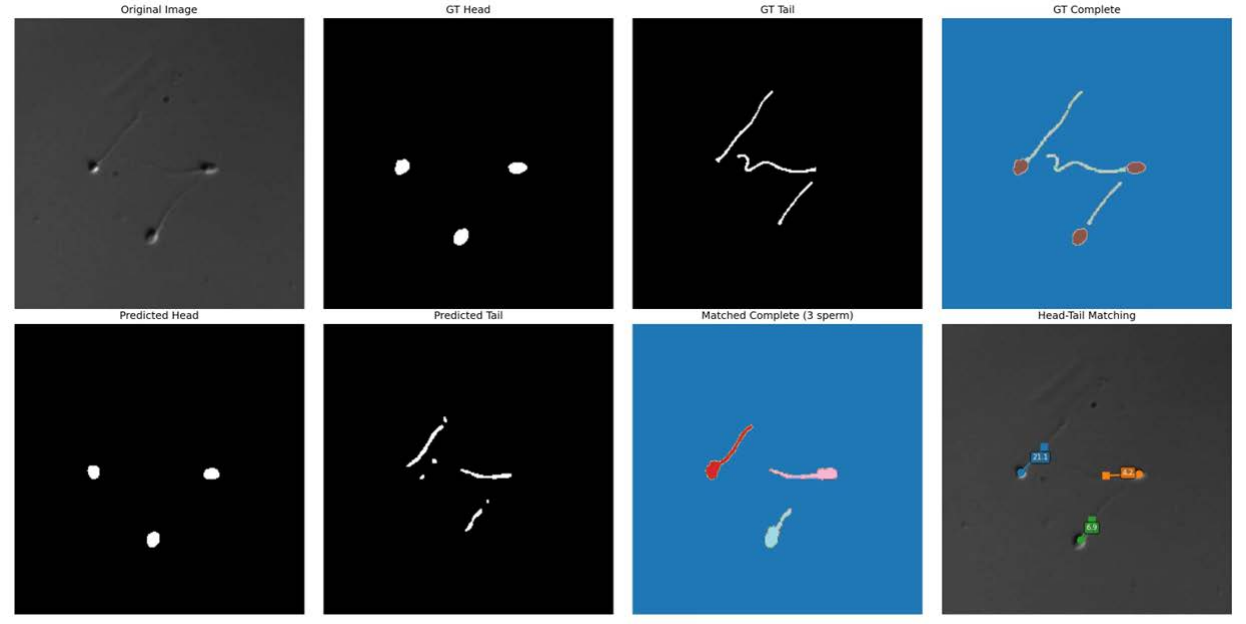}
    \vspace{-3mm}
    \caption{Representative outputs of sperm head-tail and whole-body segmentation.}
    \label{fig:seg_simple}
\end{figure*}

\begin{figure}
    \centering
    \includegraphics[width=0.49\linewidth]{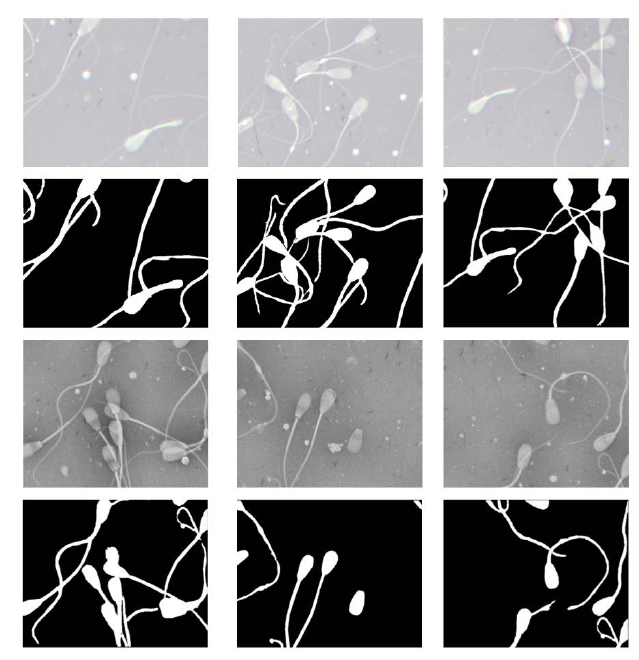}
    \vspace{-3mm}
    \caption{Instances of whole-body segmentation for sperm.}
    \label{fig:seg_hard}
\end{figure}

\subsection{Instance and Part-Level Segmentation}
\label{subsec:instance_segmentation}

Instance segmentation advances beyond semantic segmentation by simultaneously detecting individual sperm cells and generating pixel-precise masks for each instance, thereby enabling the analysis of overlapping cells and the extraction of cell-specific morphological features. Part-level segmentation further decomposes each sperm instance into its constituent biological structures: head, acrosome, nucleus, midpiece, and tail, supporting detailed morphological assessment according to strict clinical criteria.

A significant contribution to full-sperm instance segmentation was made by the FPN-integrated architecture with scale-specific cross-attention modules~\cite{lewandowska2023ensembling}, which addressed the challenge of drastic scale variations between the sperm head and tail. This work also introduced the SegSperm dataset, a curated collection of annotated sperm images that has supported subsequent research in the field. Complementing this effort, a connectivity learning framework~\cite{movahed2019automatic} was proposed to segment both external and internal sperm structures, achieving Dice scores of 0.90 for the head, 0.77 for the axoneme, and 0.75 for the tail region. By explicitly modeling structural connectivity, this approach produces anatomically coherent segmentations that respect the biological continuity of sperm components.

Part-level segmentation has also been tackled through dedicated architectures targeting specific subcellular compartments. An automated system~\cite{movahed2018learning} for segmenting the sperm head, acrosome, and nucleus achieved Dice coefficients of 0.94, 0.87, and 0.88, respectively, demonstrating that deep learning models can reliably identify boundaries between these closely apposed structures. Such granular segmentation capability is clinically relevant, as acrosomal integrity and nuclear contour regularity are established indicators of fertilization potential.

On the methodological comparison front, a comprehensive evaluation~\cite{lei2025deep} assessed Mask R-CNN, YOLOv8, YOLO11, and U-Net for multi-part sperm segmentation, providing systematic insights into the relative strengths of two-stage instance segmentation networks, single-stage unified detection-segmentation frameworks, and pure encoder-decoder architectures for this task. Additionally, Jankowski et al.~\cite{jankowski2024learning} investigated head and tail class-specific training strategies with dedicated data augmentation protocols on the SegSperm dataset, demonstrating that tailoring augmentation policies to the morphological characteristics of each sperm component can improve segmentation consistency. Figure~\ref{fig:seg_simple} and Figure~\ref{fig:seg_hard} presents exemplar results of instance-level sperm segmentation, illustrating the quality of pixel-precise delineation achieved by contemporary methods. Table~\ref{tab:segmentation_methods} provides a comprehensive overview of the segmentation techniques reviewed in this subsection.

\subsection{Foundation Models for Sperm Segmentation}
\label{subsec:foundation_models}

The emergence of foundation models: large-scale pre-trained architectures with generalized visual understanding capabilities, has opened new avenues for sperm segmentation, particularly in scenarios with limited annotated data. The Segment Anything Model (SAM)~\cite{shi2024cs3} represents a watershed development in this direction, providing a promptable segmentation framework pre-trained on an unprecedented scale of diverse imagery.

For sperm analysis, the Cascade SAM with Three Stages (CS3)~\cite{shi2024cs3} was specifically developed to adapt SAM's general-purpose segmentation capability to the sperm domain. The CS3 approach employs a cascaded prompting strategy that progressively refines segmentation masks through three stages, leveraging SAM's pre-trained visual representations while incorporating domain-specific adjustments for sperm morphology. Trained on a combination of 2,000 unlabeled and 240 expert-annotated sperm images, CS3 demonstrated that foundation models can achieve competitive segmentation performance with substantially reduced manual annotation requirements compared to training conventional architectures from scratch. This work exemplifies a broader paradigm shift in medical image analysis: rather than learning visual representations de novo from limited domain-specific data, practitioners can fine-tune or prompt pre-trained foundation models that already encode rich prior knowledge about object boundaries, textures, and shapes.

The application of foundation models to sperm segmentation holds particular promise for scenarios where large-scale expert annotation is impractical, such as rare morphological variants or specialized staining protocols. However, challenges remain in adapting these generalist models to the fine-grained structural details that distinguish normal from abnormal sperm morphology, suggesting that hybrid approaches combining foundation model priors with domain-specific refinement modules represent a promising direction for future research.

\section{Deep Learning for Sperm Morphology Classification}
\label{sec:classification}

While segmentation addresses the question of ``where''---delineating pixel-level boundaries of sperm components, morphology classification addresses the question of ``what,'' assigning semantic labels to sperm cells or their substructures based on visual features. Accurate morphological assessment is fundamental to semen analysis, as abnormal sperm morphology is strongly correlated with male infertility \cite{patel2018prediction}. Traditional manual classification following the David, Kruger (strict), or current WHO criteria is labor-intensive, subjective, and suffers from high inter-observer variability \cite{who2021laboratory,patel2018prediction,chang2017gold}. Deep learning has emerged as a powerful alternative, enabling automated, reproducible, and increasingly fine-grained classification of sperm morphological categories.

A caveat frames every result in this section. Morphology labels are themselves noisy: the ground truth is expert consensus, and the agreement among experts is often only fair. On the SCIAN gold-standard set, three-expert labelling reaches a Fleiss' $\kappa$ of only $0.36$ \cite{chang2017gold}, and inter-observer intra-class correlation for morphology can fall below $0.70$ \cite{chang2024p}. A classifier trained and evaluated against such labels is therefore measured against a noisy reference, which places an upper bound on any achievable ``accuracy'' and means that a reported $96\%$ agreement with a particular labelling protocol is not equivalent to $96\%$ agreement with biological ground truth. Accordingly, the headline percentages below should be read as agreement with a specific, imperfect annotation standard on a specific dataset, informative for relative method comparison within a benchmark, but not as evidence that models have surpassed the reliability of morphological assessment itself. This section reviews the progression from foundational CNN-based classifiers to advanced architectures such as Vision Transformers, followed by functional classification tasks including zona pellucida binding and acrosome reaction assessment, and concludes with end-to-end motility classification from video sequences. An overview of representative classification architectures is illustrated in Figure~\ref{fig:classification}.

\subsection{Convolutional Neural Networks for Morphology Classification}
\label{subsec:cnn_classification}

Convolutional neural networks established the foundation for automated sperm morphology classification by learning hierarchical visual features directly from microscopic images. A comprehensive deep learning system for sperm morphology analysis~\cite{maalej2025advancements} employed a ResNet50 architecture trained on the Sperm Morphology Dataset (SMD) and the Multi-Scanner Sperm dataset (MSS), reporting approximately 95\% classification accuracy across 12 distinct morphological defect categories and diagnosing abnormalities in each sperm component: head, midpiece, and tail, according to the David classification framework. This line of work demonstrated that CNNs can approach expert-level agreement while providing consistent, reproducible diagnoses on the benchmark used.

Subsequent research focused on enhancing CNN architectures through attention mechanisms and feature fusion to improve classification robustness. K\i l\i\c{c} \cite{kilicc2025deep} proposed a CBAM-enhanced ResNet50 architecture \cite{woo2018cbam} that integrates channel and spatial attention modules to selectively emphasize discriminative morphological features. Evaluated on both the SMIDS and HuSHeM benchmarks, their method achieved classification accuracies of 96.08\% and 96.77\%, respectively, demonstrating that attention-guided feature refinement can effectively capture subtle morphological variations distinguishing normal from abnormal sperm. In a complementary approach, a deep learning framework incorporating an Improved Bi-directional Feature Composite (IBFC) module was developed for multi-class boar sperm morphology classification \cite{keller2025deep}, reporting F1-scores ranging from 96.73\% to 99.31\% across different morphological categories and 99.8\% accuracy for acrosome health detection specifically; as a veterinary study, its class definitions are species-specific, so the architectural contribution transfers more directly than the reported operating points. These studies collectively illustrate that carefully engineered CNN architectures with attention and feature fusion mechanisms can attain near-expert performance on standardized morphological classification benchmarks.

Beyond benchmark datasets, CNN-based classification systems have been developed for practical clinical deployment under challenging imaging conditions. An automated sperm morphology analysis (SMA) algorithm \cite{ghasemian2015efficient} was designed to detect head, midpiece, and tail abnormalities directly from low-resolution unstained microscopic images, achieving over 90\% classification accuracy without requiring specialized staining protocols. This advancement is clinically significant because it enables morphological assessment during routine semen analysis without the time and cost overhead of specimen preparation. Furthermore, YOLOv3 combined with ensemble deep learning methods was applied to classify live sperm cells from real-time video streams \cite{shahali2024p,shahali2024morphology}, achieving 94\% precision and accuracy with maintained performance of 88\% under reduced spatial resolution. These practical systems underscore the translational potential of CNN-based classification for point-of-care diagnostic settings where imaging resources may be limited.

The clinical validity of automated classification has also been examined through correlation studies against expert assessments. The AI-based Open sperm Morphology (AIOM) platform \cite{chang2024p} reported intra-class correlation coefficients (ICC) exceeding $0.85$ against consensus expert evaluations, above the sub-$0.70$ inter-observer ICC of individual annotators. This pattern is consistent with a recognized property of well-regularized models, that a single consistent decision rule can align more closely with a \emph{consensus} label than any one expert whose judgments carry individual variance, rather than proof that the model exceeds expert competence, since the consensus is itself the training target. We note further that this result is currently reported as a conference abstract, so the underlying protocol, sample size, and external validation remain to be described in a full peer-reviewed account. With that caveat, these CNN-based approaches have nonetheless established automated morphological classification as a viable and increasingly reproducible component of computer-assisted semen analysis systems.

\subsection{Vision Transformers and Advanced Architectures}
\label{subsec:vit_classification}

While CNNs have demonstrated remarkable success in sperm morphology classification, the advent of Vision Transformers (ViTs) and hybrid architectures has introduced new paradigms for capturing long-range spatial dependencies and global contextual information in microscopic images. Unlike CNNs, which rely on local convolutional operations with limited receptive fields, self-attention mechanisms in transformers can model relationships between arbitrarily distant image regions, a property particularly valuable for assessing morphological coherence across the entire sperm cell.

A systematic comparative evaluation of CNN and ViT architectures for sperm morphology classification was conducted by Aktas et al. \cite{aktas2025unveiling,aktas2023performance}, who benchmarked multiple state-of-the-art models on the SMIDS and HuSHeM datasets. Their results revealed that the BEiT-Base transformer achieved superior performance with accuracies of 92.5\% on SMIDS and 93.52\% on HuSHeM, outperforming conventional CNN baselines. BEiT's pretraining strategy, which employs masked image modeling to learn robust visual representations, appears particularly well-suited for biomedical microscopy, where labeled training data is often scarce and models must generalize across varying staining protocols and imaging conditions. These findings suggest that transformer-based architectures offer distinct advantages for sperm morphology classification, particularly in scenarios requiring the integration of global shape context rather than purely local textural features.

Complementing the transformer paradigm, capsule networks have been explored as an alternative architecture capable of encoding spatial hierarchies of features through dynamic routing mechanisms. Mohammadi et al. \cite{mohammadi2024comparative} conducted a comparative study between CapsNet and an improved FixCaps architecture on the Hi-Lab sperm morphology dataset. Their experiments showed that FixCaps reached 52.67\% accuracy versus 37.89\% for the baseline CapsNet. Both figures are well below those of deeper CNN or transformer models on comparable tasks, so this comparison is best read as evidence that the FixCaps modifications help \emph{relative} to a weak capsule baseline, not as a competitive result in absolute terms. The study is nonetheless of interest because capsule architectures explicitly preserve spatial relational information between morphological parts, a property that may become more valuable as classification tasks demand finer-grained differentiation of subtle structural abnormalities, provided the absolute accuracy gap to mainstream backbones can be closed. Figure~\ref{fig:classification} summarizes the architectural diversity of classification models applied to sperm morphology analysis, spanning from foundational CNNs to attention-enhanced networks, Vision Transformers, capsule architectures, and graph-based approaches.

\subsection{Functional Classification: Zona Pellucida Binding and Acrosome Reaction}
\label{subsec:functional_classification}

Morphological normality does not guarantee functional competence; a sperm cell may appear structurally normal yet lack the physiological capacity for fertilization. Functional classification extends beyond visual morphology assessment to predict biologically meaningful capabilities---specifically, the ability to bind the zona pellucida (ZP) and undergo the acrosome reaction (AR). These functional endpoints are critical determinants of fertilization potential and are assessed in advanced semen analysis protocols as summarized in Table~\ref{tab:genetic_methods}.

The zona pellucida binding assay evaluates sperm capacity to recognize and attach to the ZP, the glycoprotein matrix surrounding the oocyte, which represents the first essential step in fertilization. Leung et al. \cite{leung2025automatic,leung2024p} developed a VGG13-based deep learning system for ZP-binding classification, training and validating their model on an extensive clinical dataset comprising over 33,000 individual sperm images. Their system achieved a sensitivity of 97.6\%, specificity of 96.0\%, and overall classification accuracy of 96.7\% for predicting ZP-binding capability from bright-field microscopy images alone (reported in a full-length study \cite{leung2025automatic} together with a companion conference abstract \cite{leung2024p}). This work is notable because it suggests that a functional, fertilization-relevant property can be predicted from standard morphology imaging without an invasive ZP-binding assay, which consumes oocytes and is restricted to specialized IVF laboratories. As with all single-centre results, external and prospective validation across devices and populations will be needed before these operating points can be assumed to hold in routine practice.

The acrosome reaction, involving the exocytosis of acrosomal enzymes necessary for ZP penetration, represents another critical functional endpoint. Park et al. \cite{park2023deep} developed a deep-learning-based system that assesses acrosome-reaction status via modification of the plasma membrane in boar spermatozoa, achieving over 97\% mean average precision (mAP) in classifying acrosome-reaction status and demonstrating that deep learning can interpret image features associated with acrosomal integrity. As a porcine study, it establishes the imaging-and-learning methodology rather than a directly transferable human operating point. Together, these functional-classification approaches begin to bridge morphological assessment and physiological competence, offering candidate non-invasive proxies for fertilization-related functional assays that would otherwise require specialized equipment and biological materials, while underscoring that human clinical validation remains the outstanding step.

\subsection{End-to-End Motility Classification from Video}
\label{subsec:motility_classification}

Beyond static morphological assessment, deep learning has been applied to classify sperm motility categories directly from video sequences, framing motility analysis as an end-to-end classification task rather than a tracking-based measurement problem. This approach leverages the spatiotemporal information contained in microscopy videos to predict clinically relevant motility grades without requiring explicit individual sperm tracking.

Hicks et al. \cite{hicks2019machine} investigated the combination of traditional machine learning with CNN features extracted from sperm movement videos, demonstrating that deep learning representations of spatiotemporal motion patterns could predict motility grades with high consistency. Their findings established that CNN-based motion analysis produces reproducible motility predictions that are less susceptible to the observer-dependent variability inherent in manual semen analysis.

Extending this approach, a deep convolutional neural network based on ResNet-50 was specifically developed for predicting WHO motility categories from sperm video recordings \cite{haugen2023sperm}. Trained and evaluated through ten-fold cross-validation on 65 semen analysis videos, the system achieved a mean absolute error (MAE) of 0.05 to 0.07 in predicting the proportions of progressive, non-progressive, and immotile sperm, with Pearson correlation coefficients of $r = 0.88$ between predicted and manually assessed motility distributions. These results indicate that end-to-end motility classification from video can provide accurate and reproducible motility assessment that aligns closely with clinical reference standards. This paradigm offers particular advantages for high-throughput screening applications, as it eliminates the computational overhead of multi-object tracking while maintaining clinically acceptable accuracy.

\begin{table}
    \centering
    \setlength\tabcolsep{2.0pt}
    \caption{AI-driven assessment of sperm functional and genetic integrity}
    \scriptsize
    \begin{tabular}{c|cccc}
    \toprule
    \rowcolor{headerblue}
       \textbf{Methods}  & \textbf{Core Idea} & \textbf{Pros} & \textbf{Cons} & \textbf{Representative} \\
       \midrule
        DNA Frag Index & DL models analyze  & Automates a labor-intensive & Requires large, annotated & \multirow{2}{*}{\cite{kumar2023deep}}\\
         (DFI) Prediction & microscopic images &  genetic assessment & datasets of halo patterns &  \\
        \midrule
        QPI for stress & Combines label-free QPI & Non-invasive, label-free, & Requires specialized & \multirow{2}{*}{\cite{butola2020high,butola2024quantitative}}  \\
        Classification & with DL models. & provides measure of sperm & QPI hardware &  \\
        \midrule
        Optical Tweezers with DL & Capture orientation and  & detailed biophysical & Specialized optical tweezers setup & \multirow{2}{*}{\cite{zhao2023deep}} \\
        for Biophysical Analysis &  rotation of sperm & analysis of sperm activity & complex experimental setup. & \\
        \midrule
        AI for Genetic & DL models analyze images (FISH) & Automates, enhances precision & Requires integration with  & \multirow{2}{*}{\cite{moustakli2024comparative,dai2021advances}} \\
        Test Interpretation & or genomic data (NGS) & of genetic assessment & complex lab techniques &  \\
    \bottomrule
    \end{tabular}
    \label{tab:genetic_methods}
\end{table}

\section{AI-Driven Analysis of Sperm Genetic Integrity and Function}
\label{sec:genetic}

While motility and morphology assessments provide foundational information for semen quality evaluation, they do not directly measure the genetic integrity or functional competence of spermatozoa. Genetic abnormalities, including DNA fragmentation, chromosomal aneuploidy, and epigenetic alterations, can compromise fertilization, embryo development, and pregnancy outcomes even when sperm appear morphologically normal and exhibit satisfactory motility \cite{liu2023integration,qu2025effect,yang2025impact}. The evaluation of sperm genetic integrity and functional status therefore represents a critical frontier in advanced assisted reproductive technology (ART). Artificial intelligence, particularly deep learning, has begun to address this frontier by developing non-invasive or minimally invasive predictors of genetic and functional parameters from imaging data. This section reviews AI-driven approaches for DNA fragmentation index (DFI) prediction, sperm stress condition and viability assessment, biophysical and genetic analysis, and their integration with microfluidic selection systems. Table~\ref{tab:genetic_methods} provides a comprehensive summary of the methods and clinical parameters addressed by these approaches.

\subsection{DNA Fragmentation Index Prediction}
\label{subsec:dfi_prediction}

Sperm DNA fragmentation refers to the presence of breaks, lesions, or structural damage in the sperm nuclear chromatin, quantified clinically as the DNA Fragmentation Index (DFI). Several studies report associations between elevated DFI and reduced fertilization rates, impaired embryo quality, increased miscarriage risk, and lower pregnancy outcomes in both natural conception and ART cycles \cite{liu2023integration,qu2025effect,yang2025impact}. It is important, however, not to overstate this evidence: the clinical utility of routine DFI testing remains contested, the association across studies is heterogeneous, assay methods are not interchangeable, and major professional-society guidance treats DFI as an adjunct rather than a standalone determinant of management \cite{eshre2025dfi}. AI-based DFI prediction should therefore be positioned as automating a specific, still-debated biomarker, not as validating that biomarker's clinical decisiveness. Conventional DFI assessment methods---including the sperm chromatin structure assay (SCSA), terminal deoxynucleotidyl transferase dUTP nick-end labeling (TUNEL), and Comet assay, require specialized fluorescent reagents, flow cytometry equipment, or labor-intensive microscopy procedures that are not routinely available in clinical andrology laboratories, which is precisely what motivates image-based surrogates.

Recent advances have demonstrated that deep learning can predict DFI directly from standard bright-field or contrast-enhanced sperm images, offering a non-invasive proxy for chromatin integrity assessment. A landmark study by Kumar et al. \cite{kumar2023deep} developed a deep learning system trained on 24,415 labeled sperm images for automated DFI prediction. Their approach framed DFI assessment as both a binary classification task (fragmented vs. non-fragmented DNA) and a multi-class categorization task across multiple DFI severity levels. The system achieved 80.15\% accuracy for binary DFI classification and 75.25\% accuracy for multi-class DFI grading. While these figures remain below the accuracy levels achieved for morphological classification tasks, they represent a clinically significant proof-of-concept: machine learning models can extract DNA integrity-related information from conventional morphology images that is not apparent to human observers. The underlying hypothesis is that subtle alterations in nuclear texture, light scattering properties, or head shape characteristics, invisible to the unaided eye but detectable by deep convolutional features, correlate with underlying chromatin structural abnormalities. These results suggest a promising pathway toward routine, non-invasive DFI screening during standard semen analysis without requiring additional specialized assays.

\subsection{Sperm Stress Condition and Viability Assessment}
\label{subsec:stress_viability}

Sperm cells in semen samples may exist in varying states of physiological stress, oxidative damage, or apoptotic progression, and these functional states can profoundly impact fertilization capacity. Distinguishing between stressed and healthy sperm populations is therefore essential for selecting the most viable gametes for ART procedures. AI-driven approaches have leveraged quantitative phase imaging (QPI) combined with deep learning to assess sperm stress conditions at the single-cell level.

Butola et al. \cite{butola2020high} developed a deep neural network (DNN) classifier operating on quantitative phase images of 10,163 individual sperm cells, achieving a sensitivity of 85.5\%, specificity of 94.7\%, and overall classification accuracy of 85.6\% for distinguishing stressed from healthy sperm populations. QPI captures biophysical cell properties, including dry mass distribution, refractive index variations, and membrane integrity markers, that encode information about cellular metabolic state and structural integrity without requiring fluorescent staining. The high specificity of this approach is clinically valuable because it minimizes the risk of falsely classifying viable sperm as non-viable, thereby preserving the pool of gametes available for fertilization. In a related development, time-delay QPI combined with deep learning was employed for temporal classification of sperm health status \cite{butola2024quantitative}, demonstrating that dynamic biophysical changes observed over time provide additional discriminative information beyond static morphological snapshots. The integration of time-resolved biophysical measurements with deep learning classifiers represents a particularly promising direction, as it captures the functional metabolic state of sperm rather than relying solely on structural appearance.

\subsection{Biophysical and Genetic Analysis}
\label{subsec:biophysical_genetic}

Beyond fragmentation and stress assessment, AI has been applied to analyze deeper functional and genetic characteristics of spermatozoa, including ion channel activity, cellular mechanics, and chromosomal integrity. These advanced analyses bridge biophysical phenotyping with genotypic assessment, offering insights into sperm functional competence at the molecular level.

CatSper, the sperm-specific calcium channel essential for hyperactivation and chemotaxis, represents a critical functional determinant that cannot be assessed through conventional morphology or motility metrics. Hwang et al. \cite{hwang2024analysis} developed an integrated approach combining conventional CASA protocols with calcium chelation experiments and machine learning analysis to evaluate CatSper functionality. Their system analyzed motility responses to calcium signaling manipulations, enabling functional characterization of this essential fertilization-related channel. The clinical relevance of such cation-channel phenotyping is underscored by reports linking sperm cation-channel dysfunction to otherwise unexplained infertility and IVF failure \cite{fayezi2025p}, which illustrates why AI methods able to interpret functional-assay data, combining behavioral phenotypes with pharmacological perturbation, are of growing interest.

At the mechanical analysis frontier, deep learning has been combined with optical tweezers for automated analysis of sperm cellular mechanics and rotational dynamics \cite{zhao2023deep}. Their system employed deep learning to automatically determine the projection orientation of ellipsoidal sperm cells trapped in optical tweezers, enabling precise analysis of sperm rotation patterns that encode information about cellular asymmetry, membrane properties, and internal structural organization. The automated orientation determination eliminated a major bottleneck in optical tweezers-based sperm analysis, which previously required labor-intensive manual alignment.

At the genetic analysis level, comparisons between fluorescence in situ hybridization (FISH) and next-generation sequencing (NGS) for sperm aneuploidy screening have informed the selection of genetic testing strategies in ART \cite{moustakli2024comparative}. While FISH provides targeted chromosomal enumeration, NGS offers comprehensive genome-wide detection of chromosomal abnormalities with higher resolution. Concurrently, advances in digital holographic microscopy, super-resolution imaging, and next-generation sequencing \cite{dai2021advances} have expanded the toolkit for correlating sperm morphological and biophysical phenotypes with underlying genetic states. These multimodal approaches, when integrated with AI-based analysis pipelines, hold promise for comprehensive sperm quality assessment that spans from molecular genetics through cellular mechanics to functional behavior.

\subsection{Integration with Microfluidic Selection}
\label{subsec:microfluidic}

Microfluidic devices have emerged as transformative platforms for sperm selection, offering precise fluidic control to isolate motile, morphologically normal, and functionally competent sperm while minimizing DNA damage. The convergence of microfluidic technology with AI-driven quality assessment creates closed-loop systems that can intelligently select optimal gametes based on integrated morphological, motility, and genetic quality criteria.

Microfluidic sperm-selection systems have been reported to improve clinical ART outcomes in specific settings \cite{naghi2025teratozoospermia}, although the strength and generality of this benefit are still under active investigation and depend on patient selection and comparator method. Mechanistically, by mimicking the natural reproductive-tract environment through precisely engineered microchannels, these devices enable passive selection of sperm with superior motility and membrane integrity, and several studies report improved fertilization rates and blastocyst quality relative to conventional density-gradient centrifugation. The gentle, biomimetic selection process minimizes oxidative stress and mechanical DNA damage that can occur during centrifugation-based sperm preparation.

The CA0 live sperm sorting device \cite{hsu2023live} exemplifies the integration of microfluidic selection with DNA integrity optimization. This system was specifically designed to minimize the DNA fragmentation index (DFI) in sorted sperm populations while enhancing overall fertilization potential. By combining microfluidic sorting with real-time quality feedback, the CA0 device achieves simultaneous selection for both motility and genetic integrity, two parameters that are not always correlated in raw semen samples. The incorporation of AI-based image analysis into such microfluidic platforms enables automated, real-time classification of sperm quality during the sorting process, dynamically adjusting selection parameters to maximize the recovery of genetically intact, functionally competent gametes. Looking forward, the deep integration of computer vision-based quality assessment with microfluidic handling is expected to enable next-generation intelligent sperm selection systems that simultaneously optimize multiple quality parameters in a fully automated workflow.

\section{Public Datasets and Benchmarks}
\label{sec:datasets}

\begin{table}[htbp]
    \centering
    \caption{Public datasets and benchmarks for AI-driven sperm analysis}
    \label{tab:sperm_datasets_optimized}
    \small
    \begin{tabularx}{\textwidth}[t]{l l l l>{\raggedright\arraybackslash}X}
    \toprule
    \rowcolor{headerblue}
    \textbf{Titles} & \textbf{Tasks} & \textbf{Images} & \textbf{Videos} & \textbf{Summary} \\
    \midrule
    \makecell[l]{HSMA-DS \cite{ghasemian2015efficient} \\ Ghasemian et al. \\ (2015)} & 
    Classification & 
    1,457 & 
    - & 
    This dataset, designed for sperm morphology analysis, comprises images captured at $\times$400 and $\times$600 magnifications. \\
    \midrule

    \makecell[l]{MHSMA \cite{javadi2019novel} \\ Soroush et al. \\ (2019)} & 
    Classification & 
    1,540 & 
    - & 
    The dataset, intended for morphology analysis, contains cropped sperm head images derived from samples of 235 participants. \\
    \midrule

    \makecell[l]{HuSHem \cite{shaker2017dictionary} \\ Shaker et al. \\ (2017)} & 
    Classification & 
    216 & 
    - & 
    HuSHem is a dataset for sperm morphology classification, comprising 131$\times$131 pixel stained sperm head images categorized into four classes: normal, tapered, pyriform, and amorphous. \\
    \midrule

    \makecell[l]{SCIAN-Morpho \cite{chang2017gold} \\ SpermGS \\ Chang et al. (2017)} & 
    Classification & 
    1,854 & 
    - & 
    This dataset supports sperm morphology analysis, providing images categorized into five classes: normal, tapered, pyriform, small, and amorphous. \\
    \midrule

    \makecell[l]{SMIDS \cite{ilhan2020fully} \\ Ilhan et al. \\ (2020)} & 
    Classification & 
    3,000 & 
    - & 
    This morphology dataset, collected from 17 subjects, includes two manually annotated classes (normal and abnormal) and one automatically extracted class (non-sperm). \\
    \midrule

    \makecell[l]{\cite{mccallum2019deep} \\ McCallum et al. \\ (2019)} & 
    Classification & 
    1064 & 
    - & 
    This dataset comprises 150 $\times$ 150 pixel bright-field sperm images from six healthy donors for selecting sperm with high DNA integrity. \\
    \midrule

    \makecell[l]{SVIA \cite{chen2022svia} \\ Chen et al. \\ (2022)} & 
    \makecell[l]{Detection, \\ Segmentation, \\ Classification} & 
    \makecell[l]{$\sim$4{,}041 \\ (+126{,}880 \\ cropped)} & 
    \makecell[l]{101 (1-2 \\ seconds)} & 
    Multi-subset dataset with $\sim$278{,}000 annotated objects: detection (3,590 full images / 125,000 object annotations), segmentation (451 images / 26,000 annotations), and classification (126,880 cropped objects). See Section~\ref{sec:datasets} for the full breakdown. \\
    \midrule

    \makecell[l]{VISEM \cite{haugen2019visem} \\ Haugen et al. \\ (2019)} & 
    Regression & 
    - & 
    85 & 
    The dataset comprises 85 videos (640$\times$480, 50 FPS) with ground truth including manually assessed semen parameters, fatty acids, sex hormones, and participant metadata. \\
    \midrule

    \makecell[l]{VISEM-Tracking \cite{thambawita2023visem} \\ Thambawita et al. \\ (2023)} & 
    \makecell[l]{Detection, \\ Tracking} & 
    \makecell[l]{29,196 \\ frames} & 
    \makecell[l]{20 (30 s) \\ +336 unlab.} & 
    Extends VISEM for tracking: 656,334 bounding-box/tracking annotations (three classes with unique IDs) over 29,196 frames at 45--50 FPS, plus 336 unannotated clips for self-/semi-supervised learning.\\
    \bottomrule
    \end{tabularx}
    \label{tab:datasets}
\end{table}

The development and validation of deep learning models for sperm analysis rely fundamentally on the availability of high-quality, annotated datasets. Public benchmarks serve as critical resources for training, comparing, and reproducing algorithms across different research groups. Table~\ref{tab:datasets} provides a comprehensive statistical summary of the publicly accessible datasets that have shaped the trajectory of AI-driven sperm analysis research. The following sections describe each dataset in detail, organized by their primary analytical focus and modality.

\textbf{SVIA Dataset} \cite{chen2022svia} represents one of the largest and most comprehensive multi-modal CASA datasets publicly available. It comprises three distinct subsets designed to address different analytical tasks: Subset-A contains 101 video sequences (MP4 format, 30 fps) with 125,000 annotated objects and 3,590 PNG images (resolution 698$\times$528$\times$3), specifically curated for sperm detection; Subset-B provides 10 video sequences encompassing 26,000 sperm ground-truth annotations and 451 images for segmentation and tracking tasks; and Subset-C offers 125,000 standalone images dedicated to denoising and classification studies. The data was acquired at JingHua Hospital using the WLJY-9000 CASA system with a 20$\times$ objective lens combined with 20$\times$ electronic magnification. Fourteen reproductive specialists contributed to the annotation process, with six senior experts performing validation, ensuring high inter-rater reliability across the dataset.

\textbf{DFI-Labeled Dataset} \cite{mccallum2019deep} constitutes the first publicly available resource linking bright-field sperm morphology with quantitative DNA quality assessment. It contains 1,064 bright-field sperm cell images collected from six healthy donors, acquired using a $\times$100 confocal microscope (Carl Zeiss Axiostar Plus) with acridine orange (AO) staining, which produces green fluorescence for double-stranded DNA (dsDNA) and red fluorescence for single-stranded DNA (ssDNA). The dataset provides sperm chromatin structure assay (SCSA) gold-standard DFI annotations with inter-expert validation, establishing a critical bridge between morphological features and molecular-level DNA integrity.

\textbf{SMIDS} (Smartphone Microscope Image Dataset for Sperm) \cite{ilhan2020fully} was specifically designed to facilitate mobile-health applications in sperm analysis. It comprises 3,000 cropped image patches distributed across three categories: 1,021 normal spermatozoa, 1,005 abnormal spermatozoa, and 974 non-sperm objects. The dataset was derived from 200 eyepiece images captured from 17 subjects using a smartphone coupled with an Olympus BX50 microscope at 20$\times$ magnification. Multi-objective genetic strategies (MOGS) denoising and Fuzzy C-Means segmentation were applied during preprocessing, making this dataset particularly suitable for resource-constrained and point-of-care diagnostic environments.

\textbf{SCIAN-MorphoSpermGS} \cite{chang2017gold} stands as the first publicly available gold-standard multi-class sperm head morphology dataset. It contains 1,854 sperm head images annotated into five morphological categories, acquired using a Carl Zeiss Axiostar Plus microscope equipped with a 63$\times$ oil immersion objective (NA 1.4) and modified hematoxylin-eosin staining. Three Chilean experts provided annotations through majority voting, achieving a Fleiss' Kappa coefficient of 0.36. This dataset has become an essential benchmark for evaluating automated sperm morphology classification systems, particularly for clinical applications requiring fine-grained morphological assessment.

\textbf{HuSHeM} (Human Sperm Head Morphology) \cite{shaker2017dictionary} provides 216 RGB images of size 131$\times$131 pixels, uniformly distributed across four morphological classes: 54 normal, 53 tapered, 57 pyriform, and 52 amorphous sperm heads. The images were collected at the Isfahan Fertility Center using an Olympus CX21 microscope at 100$\times$ magnification with Diff-Quick staining and independently verified by three domain experts. Its compact size and standardized class distribution make it particularly suitable for benchmarking classification architectures under balanced learning scenarios.

\textbf{MHSMA} (Human Sperm Morphology Analysis) \cite{javadi2019novel} contains 1,540 grayscale images at resolutions of 128$\times$128 or 64$\times$64 pixels, sourced from 235 patients. Each image is annotated across four morphological attributes: acrosome integrity, head shape, vacuole presence, and neck/tail configuration. The data was captured using an Olympus IX70 microscope at 400$\times$ or 600$\times$ magnification, with the Y channel extracted from YCbCr color space for processing. The authors demonstrated real-time applicability for intracytoplasmic sperm injection (ICSI) at a processing rate of 24 ms per image, incorporating data augmentation strategies to enhance model generalization.

\textbf{HSMA-DS} (Human Sperm Morphology Analysis Dataset) \cite{ghasemian2015efficient} offers 1,457 RGB images at 576$\times$764 resolution from 235 patients, categorized into four classes: 358 large vacuoles, 16 tail-neck defects, 502 head malformations, and 707 normal spermatozoa. Images were acquired using an Olympus IX70 microscope from unstained specimens, with annotations following Menkveld's strict criteria for morphological assessment. The notably high resolution of this dataset enables detailed sub-cellular feature extraction for fine-grained morphological analysis.

\textbf{VISEM} (Video and Image Dataset for Sperm Analysis) \cite{haugen2019visem} represents the first multi-modal dataset integrating sperm analysis data with biological and demographic information. It encompasses data from 85 participants, including 85+ AVI video recordings of semen samples, accompanied by fatty acid profiles, sex hormone levels, and comprehensive demographic metadata. All video data was captured using an Olympus CX31 microscope at 400$\times$ magnification with a 37$^\circ$C heated stage to maintain physiological motility conditions. This multi-modal structure enables investigations into the relationships between semen quality parameters and broader physiological and lifestyle factors.

\textbf{VISEM-Tracking} \cite{thambawita2023visem} extends the VISEM dataset specifically for object tracking research. It provides 20 thirty-second video sequences containing 29,196 frames captured at 45--50 FPS, with 656,334 bounding box annotations across three semantic categories, each assigned unique tracking IDs. Additionally, 336 unannotated thirty-second video clips are included for semi-supervised and self-supervised learning paradigms. This dataset has become the primary benchmark for evaluating multi-object tracking algorithms in the context of sperm motility analysis.

Collectively, these datasets address diverse aspects of sperm analysis, from morphological classification and DNA integrity assessment to motility tracking and multi-modal reproductive health profiling. Their availability has been instrumental in establishing evaluation protocols and accelerating the translation of deep learning research into practice. Three structural limitations, however, temper how far current benchmarks can support claims of generalizable clinical performance. First, most sets are small (hundreds to a few thousand images) and single-centre, which inflates optimistic in-distribution scores and limits external validity. Second, they are heterogeneous in imaging modality and preparation: bright-field, phase-contrast, and confocal acquisition; stained (Papanicolaou, Diff-Quik, hematoxylin--eosin) versus unstained specimens; and differing magnifications, so that a percentage reported on one set is not commensurable with the same metric on another. Third, and most consequential for the field's fairness and robustness ambitions, these datasets almost never report the demographic composition (age, ethnicity, geography, infertility etiology) of the men from whom samples were drawn. This omission directly undercuts the population-level generalization arguments made in Section~\ref{sec:robustness}: one cannot demonstrate cross-population robustness on benchmarks that do not disclose population. We therefore recommend that future dataset releases include a documented ``population representativeness'' field alongside standardized, modality-annotated splits and multi-centre collection, so that reported performance can be interpreted, and audited, against the population it actually reflects.

\section{Evaluation Metrics for Sperm Analysis}
\label{sec:metrics}

The quantitative evaluation of deep learning models for sperm analysis necessitates task-specific metrics that capture different facets of algorithmic performance, from individual classification accuracy to spatial localization precision and temporal tracking consistency. This section provides a systematic review of the principal evaluation metrics employed across four major task categories: classification, object detection, semantic segmentation, and multi-object tracking. Table~\ref{tab:metrics} summarizes the key metrics and their respective applications. Where applicable, we present formal mathematical definitions and discuss the relevance of each metric to specific challenges in sperm analysis.

\begin{table}[htbp]
    \centering
    \caption{Evaluation metrics for sperm analysis tasks}
    \label{tab:evaluation_metrics_vline}
    \small
    \begin{tabularx}{0.8\textwidth}{l | >{\raggedright\arraybackslash}X}
    \toprule
    \rowcolor{headerblue}
    \textbf{Tasks} & \textbf{Metrics} \\
    \midrule
    Classification & Accuracy, Precision, Recall, F1-score, ROC-AUC \\
    \midrule
    Detection & IoU, AP, mAP, PRC \\
    \midrule
    Segmentation & Pixel Accuracy, mPA, IoU, mIoU, Dice Coefficient \\
    \midrule
    Tracking & MOTA, MOTP, HOTA, ID Switches, Fragmentation, IDF1 \\
    \bottomrule
    \end{tabularx}
    \label{tab:metrics}
\end{table}

\subsection{Classification Metrics}
\label{subsec:classification_metrics}

Classification tasks in sperm analysis, such as normal versus abnormal sperm categorization, DNA fragmentation assessment, and morphological typing, rely on metrics derived from the confusion matrix. The following metrics quantify different aspects of classification performance:

\textbf{Accuracy} (ACC) measures the proportion of correctly classified instances among all predictions:
\begin{equation}
\label{eq:accuracy}
\text{ACC} = \frac{TP + TN}{TP + TN + FP + FN}
\end{equation}
where $TP$, $TN$, $FP$, and $FN$ denote true positives, true negatives, false positives, and false negatives, respectively. While widely adopted, accuracy can be misleading in imbalanced datasets, a common scenario in clinical sperm analysis where normal samples typically outnumber abnormal cases.

\textbf{Precision} quantifies the reliability of positive predictions, representing the fraction of correctly identified positive instances:
\begin{equation}
\label{eq:precision}
\text{Precision} = \frac{TP}{TP + FP}
\end{equation}
In sperm analysis, high precision is critical for clinical applications where false positives may lead to unnecessary medical interventions or inappropriate treatment decisions.

\textbf{Recall} (also termed sensitivity) measures the completeness of positive instance detection:
\begin{equation}
\label{eq:recall}
\text{Recall} = \frac{TP}{TP + FN}
\end{equation}
High recall ensures that pathological sperm phenotypes, such as morphological abnormalities or compromised DNA integrity, are not overlooked during clinical screening, which is essential for reliable male fertility assessment.

\textbf{F1-score} provides the harmonic mean of precision and recall, offering a balanced measure when class distributions are uneven:
\begin{equation}
\label{eq:f1}
F_1 = 2 \cdot \frac{\text{Precision} \cdot \text{Recall}}{\text{Precision} + \text{Recall}} = \frac{2 \cdot TP}{2 \cdot TP + FP + FN}
\end{equation}
The F1-score is particularly valuable for sperm morphology classification tasks where imbalanced class distributions are prevalent and neither precision nor recall alone adequately captures clinical utility.

Additionally, the \textbf{Receiver Operating Characteristic--Area Under Curve} (ROC-AUC) assesses model discrimination across all classification thresholds, providing a threshold-independent performance summary. It is especially useful for clinical decision support systems where operating points may be adjusted based on clinical priorities.

\subsection{Object Detection Metrics}
\label{subsec:detection_metrics}

Object detection in sperm analysis requires simultaneous localization and classification of individual sperm cells in microscopy images or video frames. The following metrics evaluate spatial detection accuracy:

\textbf{Intersection over Union} (IoU), also known as the Jaccard index for bounding boxes, quantifies the spatial overlap between predicted and ground-truth bounding boxes:
\begin{equation}
\label{eq:iou}
\text{IoU} = \frac{A_{\text{pred}} \cap A_{\text{gt}}}{A_{\text{pred}} \cup A_{\text{gt}}}
\end{equation}
where $A_{\text{pred}}$ and $A_{\text{gt}}$ denote the areas of the predicted and ground-truth bounding boxes, respectively. IoU values range from 0 (no overlap) to 1 (perfect overlap), with a threshold of 0.5 typically adopted for positive detection assignment in sperm detection benchmarks.

\textbf{Average Precision} (AP) computes the area under the precision-recall curve across varying confidence thresholds, summarizing detection performance for a single class:
\begin{equation}
\label{eq:ap}
\text{AP} = \int_0^1 p(r) \, dr
\end{equation}
where $p(r)$ represents the precision at recall level $r$. AP integrates both the accuracy of detections (through precision) and the coverage of ground-truth objects (through recall), making it a robust single-value summary for detection quality.

\textbf{Mean Average Precision} (mAP) extends AP across all object categories, providing a unified detection performance measure:
\begin{equation}
\label{eq:map}
\text{mAP} = \frac{1}{C} \sum_{c=1}^{C} \text{AP}_c
\end{equation}
where $C$ is the number of semantic classes and $\text{AP}_c$ denotes the average precision for class $c$. In sperm analysis, mAP is typically computed across sperm and non-sperm categories, or across different morphological subtypes when fine-grained detection is required.

The \textbf{Precision-Recall Curve} (PRC) additionally provides a visual representation of the trade-off between precision and recall at varying detection thresholds, offering insights into the operating characteristics of detection models across different clinical sensitivity requirements.

\subsection{Semantic Segmentation Metrics}
\label{subsec:segmentation_metrics}

Semantic segmentation in sperm analysis enables pixel-level delineation of sperm anatomical structures, including the head, acrosome, nucleus, and flagellum. The following metrics evaluate segmentation fidelity:

\textbf{Pixel Accuracy} (PA) computes the proportion of correctly classified pixels:
\begin{equation}
\label{eq:pa}
\text{PA} = \frac{\sum_{i=1}^{C} n_{ii}}{\sum_{i=1}^{C} \sum_{j=1}^{C} n_{ij}}
\end{equation}
where $n_{ij}$ represents the number of pixels belonging to class $i$ that are predicted as class $j$, and $C$ is the total number of semantic classes. While intuitive, PA can be dominated by large background regions in microscopy images.

\textbf{Mean Pixel Accuracy} (mPA) addresses this limitation by averaging per-class pixel accuracies, ensuring balanced evaluation across anatomical structures of varying sizes.

\textbf{Mean Intersection over Union} (mIoU), the most widely adopted segmentation metric, computes the average IoU across all semantic classes:
\begin{equation}
\label{eq:miou}
\text{mIoU} = \frac{1}{C} \sum_{i=1}^{C} \frac{n_{ii}}{\sum_{j=1}^{C} n_{ij} + \sum_{j=1}^{C} n_{ji} - n_{ii}}
\end{equation}
mIoU penalizes both false positives and false negatives for each class, providing a rigorous assessment of segmentation quality that is particularly sensitive to boundary delineation accuracy in sperm morphology analysis.

\textbf{Dice Coefficient}, also known as the F1-score for segmentation, measures the spatial overlap between predicted and ground-truth segmentations:
\begin{equation}
\label{eq:dice}
\text{Dice} = \frac{2 \cdot |A_{\text{pred}} \cap A_{\text{gt}}|}{|A_{\text{pred}}| + |A_{\text{gt}}|} = \frac{2 \cdot \sum_{i=1}^{C} n_{ii}}{\sum_{i=1}^{C} \sum_{j=1}^{C} n_{ij} + \sum_{j=1}^{C} \sum_{i=1}^{C} n_{ij}}
\end{equation}
The Dice coefficient is particularly prevalent in medical image segmentation literature due to its favorable gradient properties during training and its interpretable range of $[0, 1]$, making it well-suited for sperm structure delineation tasks.

\subsection{Multi-Object Tracking Metrics}
\label{subsec:tracking_metrics}

Multi-object tracking (MOT) evaluation in sperm analysis presents unique challenges due to the high density, morphological similarity, and frequent occlusions among sperm cells in microscopy video. Standardized metrics from the MOT community have been adapted for sperm tracking benchmarks:

\textbf{Multi-Object Tracking Accuracy} (MOTA) measures tracking accuracy by aggregating three error sources: false positives, false negatives, and identity switches, normalized by the total number of ground-truth objects:
\begin{equation}
\label{eq:mota}
\text{MOTA} = 1 - \frac{\sum_t (FN_t + FP_t + IDSw_t)}{\sum_t GT_t}
\end{equation}
where $FN_t$, $FP_t$, and $IDSw_t$ denote false negatives, false positives, and identity switches at frame $t$, respectively, and $GT_t$ is the number of ground-truth objects in frame $t$. MOTA provides a comprehensive tracking accuracy measure but can be dominated by detection errors rather than association quality.

\textbf{Multi-Object Tracking Precision} (MOTP) evaluates the localization precision of tracked objects:
\begin{equation}
\label{eq:motp}
\text{MOTP} = \frac{\sum_{i,t} d_{i,t}}{\sum_t c_t}
\end{equation}
where $d_{i,t}$ represents the bounding box IoU between the $i$-th matched detection-track pair at frame $t$, and $c_t$ is the number of matches at frame $t$. MOTP reflects the spatial accuracy of successful track-to-detection associations.

\textbf{Identity Switches} (IDSw) quantifies the number of times a tracker incorrectly swaps the identities of two tracked objects. Minimizing IDSw is particularly critical in sperm motility analysis, where maintaining consistent identity assignment across frames enables accurate computation of kinematic parameters such as curvilinear velocity (VCL) and straight-line velocity (VSL).

\textbf{Fragmentation} (Frag) counts the number of times a ground-truth trajectory is interrupted and later resumed. High fragmentation indicates temporally inconsistent tracking, which can adversely affect motility pattern analysis and behavioral classification.

\textbf{IDF1} measures the ratio of correctly identified detections over the average number of ground-truth and predicted detections, emphasizing identity preservation:
\begin{equation}
\label{eq:idf1}
\text{IDF1} = \frac{2 \cdot \sum_{i,t} \mathbb{1}(\hat{a}_{i,t} = a_{i,t})}{\sum_t GT_t + \sum_t DT_t}
\end{equation}
where $\hat{a}_{i,t}$ and $a_{i,t}$ denote the predicted and true identity assignments for detection $i$ at frame $t$, and $DT_t$ is the number of detections at frame $t$. IDF1 has gained prominence in sperm tracking evaluation due to its sensitivity to identity consistency, which is paramount for reliable long-term motility profiling.

\textbf{Higher Order Tracking Accuracy} (HOTA) \cite{luiten2021hota} was introduced to address a known deficiency of MOTA, namely its tendency to be dominated by detection error rather than association quality. HOTA decomposes tracking performance into an explicit balance of detection accuracy and association accuracy, integrated over a range of localization thresholds, and correlates better with human judgments of tracking quality. It is increasingly reported in sperm tracking studies alongside MOTA and IDF1 and is preferable when the goal is to characterize identity preservation, which is the property that actually governs kinematic-parameter reliability.

\smallskip
\noindent\textbf{Reporting pitfalls specific to sperm analysis.} Beyond their definitions, these metrics are frequently misused in ways that inflate apparent performance. (i) \emph{Accuracy on imbalanced data}: because normal cells and background dominate typical fields, high accuracy can coexist with poor detection of the rare abnormal or pathological cases that matter clinically; balanced accuracy, per-class recall, and F1 should accompany accuracy. (ii) \emph{Under-specified detection thresholds}: an mAP figure is uninterpretable without stating the IoU threshold(s) (e.g., AP$_{50}$ versus AP$_{50:95}$), yet these are often omitted, making cross-study numbers incomparable. (iii) \emph{MOTA as a proxy for motility quality}: since MOTA is detection-dominated, a high MOTA can mask the identity switches and fragmentation that corrupt kinematic estimates; HOTA, IDF1, IDSw, and Frag should therefore be reported together for any motility-oriented tracker. (iv) \emph{Metric--endpoint mismatch}: image-level scores (mAP, Dice) are necessary but not sufficient evidence of clinical utility, which is defined at the sample and patient level; we return to this hierarchy in Section~\ref{sec:robustness}. Reporting these details is inexpensive and is a prerequisite for the cross-centre comparability the field currently lacks.

\section{Emerging Applications and Advanced Deep Learning Paradigms}
\label{sec:emerging}

While convolutional neural networks and standard supervised learning pipelines have demonstrated remarkable success in sperm analysis, the field is rapidly evolving toward more sophisticated paradigms that address persistent challenges including data scarcity, model generalization, clinical interpretability, and multi-modal integration. This section examines three emerging directions: multimodal fusion, strategies for data scarcity and generalization, and explainable AI with clinical integration—that may inform future intelligent sperm analysis systems if validated in clinically realistic settings. Table~\ref{tab:paradigms} provides a comparative overview of these advanced paradigms and their representative methodologies.

\renewcommand\arraystretch{1.25}
\newcolumntype{M}[1]{>{\centering\arraybackslash}m{#1}}
\newcolumntype{P}[1]{>{\raggedright\arraybackslash}m{#1}} 

\begin{table}[htbp]
    \centering
    \caption{Clinical translation challenges and mitigation strategies}
    \label{tab:advanced_paradigms}
    \small
    \resizebox{\columnwidth}{!}{%
    \begin{tabular}{P{2.8cm} P{4.2cm} P{4.5cm} P{4.5cm} M{3.0cm}}  
    \toprule
    \rowcolor{headerblue}
    \textbf{Paradigms} & \textbf{Core Idea} & \textbf{Advantages for Sperm Analysis} & \textbf{Disadvantages/Challenges} & \textbf{Representative Papers} \\
    \midrule

    \makecell[c]{Multimodal Data \\ Fusion} & 
    Combines information from different data types (e.g., images, videos, clinical data, genetic markers). &
    More comprehensive and accurate diagnoses, holistic assessment of sperm quality. &
    Data heterogeneity, complex model architectures, challenges in feature alignment. &
    \cite{goh2024multimodal} \\
    \midrule

    \makecell[c]{Self-Supervised \\ Learning (SSL)} &
    Models learn representations from unlabeled data by solving pretext tasks. &
    Reduces reliance on extensive manual annotations, leverages vast amounts of unlabeled data. &
    Pretext task design is crucial, may not capture all nuances of specific medical tasks. &
    \makecell[c]{\cite{zhuang2025mim,zhang2023dive,haghighi2024self,zhou2023unified}} \\
    \midrule

    \makecell[c]{Federated \\ Learning (FL)} &
    Collaborative model training across multiple institutions without sharing raw data. &
    Addresses data privacy concerns, enables model generalization across diverse populations/clinics. &
    Communication overhead, handling non-IID data distributions, potential for model drift. &
    \makecell[c]{\cite{che2025feddag,wu2024facmic,xie2024mh,wang2024testing}} \\
    \midrule

    \makecell[c]{Explainable AI \\ (XAI)} &
    Techniques to make deep learning models more transparent and interpretable. &
    Builds trust among clinicians, aids in clinical decision-making, helps identify model biases. &
    Can be complex to implement, trade-off between interpretability and accuracy, subjective evaluation of explanations. &
    \makecell[c]{\cite{ma2023implementation,hao2023towards,hao2023deep}} \\
    \midrule

    \makecell[c]{Generative Adversarial \\ Networks (GANs) for \\ Data Augmentation} &
    Generates synthetic, realistic images to expand training datasets. &
    Mitigates data scarcity, improves model robustness and generalization. &
    Potential for generating unrealistic or biased samples, quality control of synthetic data. &
    \cite{abbasi2023transfer} \\
    \bottomrule
    \end{tabular}
    }
    \label{tab:paradigms}
\end{table}

\subsection{Multimodal Fusion for Integrated Sperm Assessment}
\label{subsec:multimodal_fusion}

\begin{figure}
    \centering
    \includegraphics[width=0.998\linewidth]{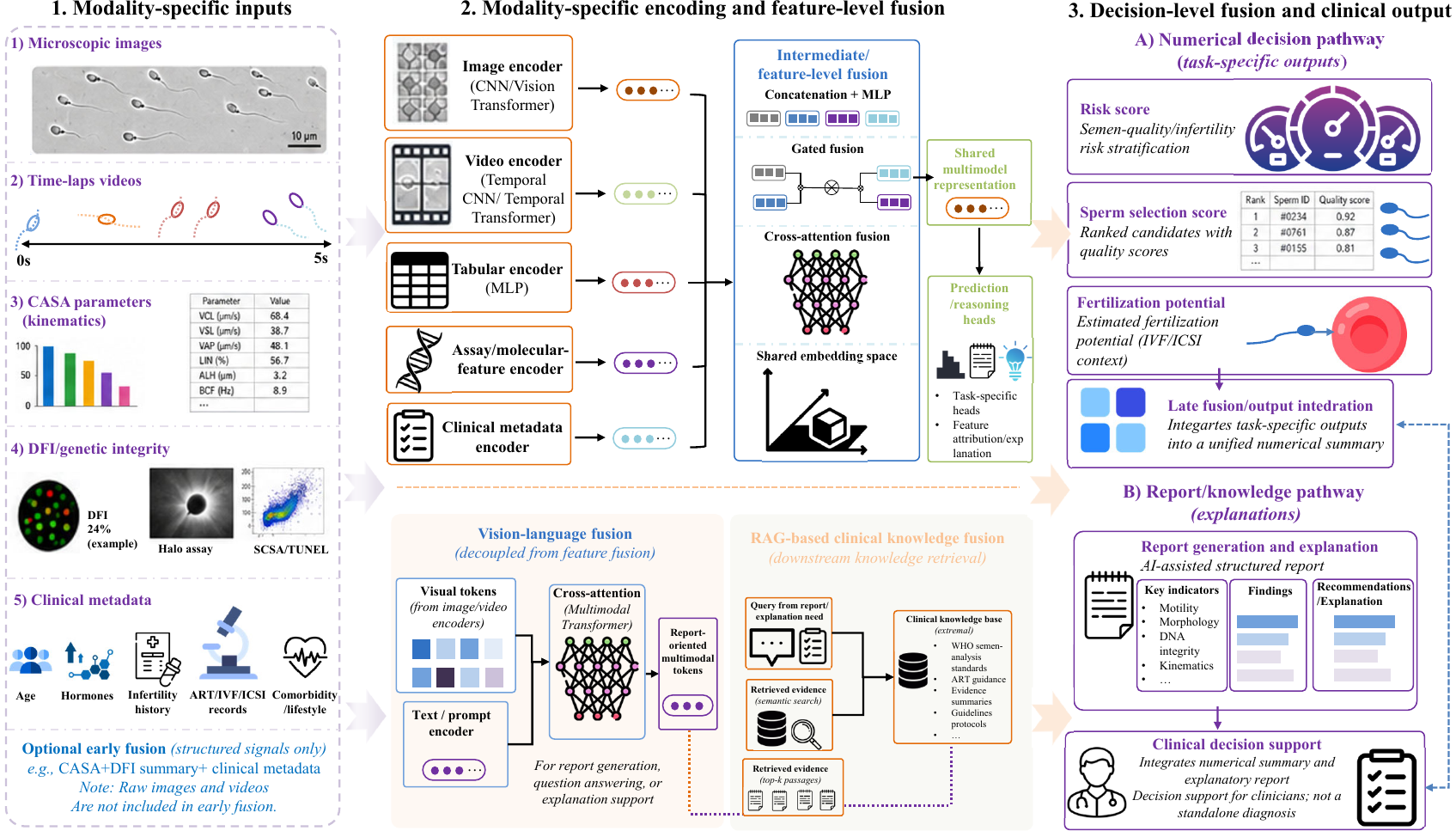}
    \vspace{-4mm}
    \caption{The overview of the fusion paradigm of multi-source data regarding the sperm characteristics.}
    \label{fig:fusion_paradigm}
\end{figure}

\noindent\textbf{Problem formulation.} To frame sperm assessment explicitly as an information-fusion problem, let a sample be described by a set of $M$ heterogeneous modalities $\{\mathbf{x}_1,\dots,\mathbf{x}_M\}$, where $\mathbf{x}_m$ may denote a microscopic image, a time-lapse motility video, a vector of CASA-derived kinematic parameters, a DNA-integrity (DFI) assay readout, or structured clinical metadata. Each modality is mapped to a latent representation by a modality-specific encoder, a fusion operator $\Phi$ produces a joint representation, and one or more task heads $g_k$ predict clinically relevant endpoints $\hat{y}_k$ (e.g., motility grade, morphology class, DFI, sperm-selection score, or ART outcome):
\begin{equation}
\label{eq:fusion_formulation}
\mathbf{z}_m = f_m(\mathbf{x}_m;\theta_m),
\qquad
\mathbf{z} = \Phi(\mathbf{z}_1,\dots,\mathbf{z}_M),
\qquad
\hat{y}_k = g_k(\mathbf{z}).
\end{equation}
Within this formulation, fusion strategies are commonly categorized as \emph{early} (input/data-level), \emph{intermediate} (feature-level, i.e., $\Phi$ acts on $\{\mathbf{z}_m\}$), and \emph{late} (decision-level, i.e., $\Phi$ aggregates the modality-specific predictions $\{\hat{y}_k^{(m)}\}$). Two properties are essential in the clinical setting and recur throughout this section: (i) \emph{missing-modality robustness}, since real datasets rarely contain synchronized image, video, assay, and outcome data for the same patient, so that $\Phi$ must operate over an arbitrary available subset $\mathcal{S}\subseteq\{1,\dots,M\}$,
\begin{equation}
\label{eq:fusion_missing}
\mathbf{z} = \Phi\big(\{\mathbf{z}_m\}_{m\in\mathcal{S}}\big),
\qquad
\mathcal{S}\subseteq\{1,\dots,M\};
\end{equation}
and (ii) \emph{uncertainty-aware weighting}, in which each modality contributes according to an estimated reliability $w_m$ so that noisy or low-quality modalities do not dominate the joint prediction,
\begin{equation}
\label{eq:fusion_weighting}
\mathbf{z} = \Phi\big(\{w_m\,\mathbf{z}_m\}_{m\in\mathcal{S}}\big),
\qquad
w_m \ge 0.
\end{equation}
Table~\ref{tab:fusion_methods} summarizes the fusion levels against these criteria and the current evidence status in sperm analysis.

\begin{table}[htbp]
    \centering
    \setlength\tabcolsep{2.0pt}
    \caption{Fusion strategies for AI-driven sperm analysis, characterized by fusion level, suitable modality types, handling of missing modalities and uncertainty, and current evidence status in andrology. ``Direct'' = evaluated on sperm/semen data; ``Transferable'' = established in broader biomedical AI but not yet validated for sperm; ``Prospective'' = conceptual direction requiring dedicated validation.}
    \label{tab:fusion_methods}
    \scriptsize
    \begin{tabularx}{\linewidth}{l >{\raggedright\arraybackslash}X >{\raggedright\arraybackslash}X c}
    \toprule
    \rowcolor{headerblue}
    \textbf{Fusion level} & \textbf{Best-suited modalities} & \textbf{Missing-modality / uncertainty handling} & \textbf{Evidence} \\
    \midrule
    Early (data-level) & Compatible structured signals: CASA parameters, DFI summaries, hormonal profiles, patient metadata & Poor: requires synchronized, complete inputs; brittle to missing data & Direct \\
    \midrule
    Intermediate (feature-level) & Image/video/tabular/assay embeddings via modality-specific encoders; cross-attention or gated fusion & Moderate: gating and attention can down-weight modalities; supports partial subsets $\mathcal{S}$ & Transferable \\
    \midrule
    Late (decision-level) & Task-specific heads: motility, morphology, DFI, selection, risk scores & Good: robust to missing modalities; naturally supports uncertainty-weighted ensembling & Direct \\
    \midrule
    Vision--language (decoupled) & Visual tokens + clinical text/report prompts for report generation and QA & N/A for numeric prediction; used for explanation, not primary scoring & Transferable \\
    \midrule
    Retrieval-augmented (RAG) & External guidelines/standards retrieved to ground clinician-facing outputs & Knowledge support only; must not modify numeric predictions without validation & Prospective \\
    \bottomrule
    \end{tabularx}
\end{table}

Sperm quality assessment is inherently multimodal, involving static morphology, dynamic motility, quantitative CASA parameters, DNA integrity assays, and patient-level clinical information. Conventional semen analysis already relies on multiple biological and technical signals, including sperm concentration, motility, morphology, and laboratory-standardized interpretation according to semen-analysis criteria \cite{patel2018prediction,you2021machine,dai2021advances}. Recent AI studies have further expanded this scope to include image-based morphology analysis, video-based motility assessment, functional sperm evaluation, DNA fragmentation prediction, and label-free optical assessment \cite{abou2023artificial,cherouveim2023artificial,kumar2023deep,butola2020high,butola2024quantitative,leung2025automatic,park2023deep}. However, the current evidence base for multimodal AI in sperm analysis remains relatively limited compared with single-task computer vision models for detection, tracking, segmentation, and morphology classification. Therefore, multimodal fusion should be discussed as an emerging methodological direction rather than as a clinically established solution. In this review, we distinguish between three levels of evidence: methods directly evaluated in sperm analysis, transferable strategies developed in broader biomedical AI, and prospective directions that require dedicated validation in andrology and ART settings.

A technically sound multimodal framework should avoid naive fusion of heterogeneous raw modalities. Microscopic images, time-lapse videos, CASA-derived kinematic summaries, DNA fragmentation assays, and clinical metadata differ substantially in data structure, temporal resolution, noise characteristics, and biological meaning. Multimodal datasets such as VISEM illustrate the value of combining semen videos with clinical and biological variables, but they also highlight the practical complexity of aligning heterogeneous sperm-related data sources \cite{haugen2019visem,thambawita2023visem}. Therefore, raw images and videos should generally undergo modality-specific preprocessing and encoding before being integrated with structured laboratory or clinical variables. Early fusion is most appropriate for compatible structured signals, such as CASA parameters, DFI summaries, hormonal profiles, or patient metadata. In contrast, feature-level fusion can combine modality-specific embeddings generated by image encoders, video encoders, tabular encoders, assay-specific encoders, and clinical embedding layers. Decision-level fusion may then integrate outputs from task-specific heads, such as motility scores, morphology classifications, DNA integrity estimates, sperm selection scores, and sample-level risk predictions.

Direct evidence for multimodal learning in sperm analysis is still emerging. A representative study employed multimodal convolutional neural networks by integrating image-based and video-based inputs, using 3D CNNs for motility-related feature extraction and 2D CNN or ResNet-based backbones for concentration prediction \cite{goh2024multimodal}. This multi-stream design provides an important proof of concept: static morphology and dynamic motion encode different aspects of sperm quality and may benefit from joint representation learning. Nevertheless, most available studies remain limited to restricted modality combinations, relatively narrow endpoints, and retrospective evaluation settings. Robust evidence for integrating microscopic imaging, time-lapse motility videos, CASA parameters, DNA integrity assays, clinical metadata, and ART outcomes within a single externally validated system is still lacking.

From an information-fusion perspective, future work should focus less on simply adding more modalities and more on determining which modalities are clinically complementary, temporally aligned, and reliable under real-world laboratory conditions. For example, morphology classification models can capture structural abnormalities \cite{shahali2024morphology,maalej2025advancements,kilicc2025deep,aktas2025unveiling}, tracking and end-to-end video models can quantify or predict motility-related behavior \cite{somasundaram2021faster,zhang2024sperm,haugen2023sperm}, DNA fragmentation models can support automated integrity assessment \cite{kumar2023deep,mccallum2019deep}, and QPI-based models may provide non-invasive functional information \cite{butola2020high,butola2024quantitative}. These modalities may contain non-redundant information, but their incremental value should be quantified through ablation studies, missing-modality testing, calibration analysis, and external validation. Fusion models should also explicitly handle incomplete data, because clinical datasets often lack synchronized images, videos, molecular assays, and longitudinal ART outcomes for the same patient or sample. Modality weighting, uncertainty-aware fusion, gated fusion, and late-fusion ensembles may therefore be more practical than fully end-to-end architectures in early clinical translation.

Vision-language fusion should be positioned separately from conventional feature-level fusion for numerical prediction. In sperm analysis, visual encoders may transform microscopic images or motility videos into visual tokens, while text encoders may process report templates, clinical prompts, or structured descriptions. In broader biomedical AI, vision-language and generalist biomedical models have shown the feasibility of aligning medical images and text for segmentation, classification, question answering, or report-oriented reasoning \cite{koleilat2024medclip,yu2024core,Singhal2023MedPaLM,Singhal2023MedPaLM2,Tu2023MedPaLMM,Zhang2023BiomedGPT,Liu2023MedicalMLLM,Yan2023GPT4VMedical,Lu2024PathChat}. However, direct evidence for vision-language models in sperm analysis remains limited. Therefore, these models should currently be regarded as transferable paradigms from broader biomedical AI rather than validated diagnostic tools for andrology. Their potential value lies primarily in improving interpretability, documentation, and human-AI interaction, provided that generated outputs are constrained by clinical evidence, uncertainty estimates, and expert oversight.

Retrieval-augmented generation \cite{lewis2020retrieval} should be considered a downstream knowledge-support mechanism rather than a core feature-fusion module. In a future sperm-analysis system, retrieval could be used to obtain relevant passages from external clinical knowledge bases, such as semen-analysis standards, ART guidance, evidence summaries, or laboratory protocols, to support report generation and explanation. For example, when a model identifies abnormal motility or elevated DNA fragmentation, a retrieval module could help ground the generated explanation in accepted terminology and guideline-consistent interpretation \cite{patel2018prediction,dai2021advances}. Importantly, retrieved knowledge should not be treated as an additional raw patient modality, nor should it directly modify numerical predictions without validation. Its appropriate role is to improve transparency, traceability, and guideline alignment of clinician-facing outputs.

Multi-agent clinical reasoning should be framed as a longer-term conceptual direction rather than an immediately deployable framework for sperm analysis. Recent surveys suggest that LLM-based multi-agent systems may support complex problem solving, planning, and critique-based reasoning \cite{Guo2024MultiAgentSurvey}. In principle, specialized agents could be assigned to imaging interpretation, motility analysis, genetic-integrity assessment, guideline retrieval, safety checking, and clinician-facing explanation. Such decomposition is conceptually attractive because infertility management is inherently multidisciplinary. However, no mature multi-agent system has yet been clinically validated for sperm analysis or ART decision-making. Moreover, agent-based systems introduce additional risks, including error propagation, overconfident consensus, inconsistent reasoning among agents, unclear accountability, and difficulty in auditing intermediate decisions. Therefore, multi-agent architectures should be discussed only as a future research hypothesis that requires evidence grounding, uncertainty estimation, explicit human oversight, and prospective clinical evaluation before any reproductive-medicine deployment.

Overall, multimodal fusion is best viewed as a promising but still under-validated direction for AI-driven sperm analysis. Its clinical value will depend not only on architectural sophistication but also on data quality, modality alignment, missing-data robustness, uncertainty calibration, and linkage to clinically meaningful endpoints. Reporting, evaluation, and regulatory frameworks further emphasize risk management, data quality, transparency, human oversight, model-update control, and post-market monitoring for AI systems used in medical contexts \cite{Collins2024TRIPODAI,Vasey2022DECIDEAI,Jin2022ClinicalXAI,FDA2024PCCP,EuropeanUnion2024AIAct}. The next stage of research should therefore move from proof-of-concept fusion models toward multicenter datasets that connect sperm images, motility videos, CASA parameters, DNA integrity assays, patient-level variables, and ART outcomes. Only through such validation can multimodal sperm AI progress from an attractive information-fusion framework to trustworthy clinical decision support.

\subsection{Addressing Data Scarcity and Generalization}
\label{subsec:data_scarcity}

The acquisition and expert annotation of large-scale sperm microscopy datasets remain prohibitively expensive and time-consuming, creating a persistent bottleneck for supervised learning approaches. Three complementary paradigms: self-supervised learning (SSL), federated learning (FL), and generative data augmentation, have emerged as powerful strategies to mitigate data scarcity.

\textbf{Self-Supervised Learning} enables models to learn meaningful representations from unlabeled data through pretext tasks. Masked image modeling (MiM) \cite{zhuang2025mim} has shown particular promise for medical imaging, where models learn to reconstruct masked image patches, acquiring anatomical understanding without manual annotations. PCRLv2 \cite{zhou2023unified} advances this approach through patch-level contrastive learning that preserves spatial relationships, while DiRA \cite{haghighi2024self} introduces disentangled representation learning that separates domain-invariant semantic features from domain-specific characteristics. Large-scale empirical studies \cite{zhang2023dive} have further validated that self-supervised pre-training on large unlabeled medical image corpora consistently improves downstream task performance, even with limited annotated data.

\textbf{Federated Learning} enables collaborative model training across decentralized clinical institutions without sharing sensitive patient data. FedDAG \cite{che2025feddag} addresses the domain heterogeneity challenge inherent in multi-center sperm analysis through domain aggregation and generalization strategies. FACMIC \cite{wu2024facmic} combines federated learning with foundation model knowledge distillation, enabling resource-constrained laboratories to benefit from large-scale pre-trained representations. MH-pFLGB \cite{xie2024mh} introduces personalized federated learning with gradient-based aggregation, which accommodates the statistical heterogeneity across different clinical populations. The integration of diverse training data from multiple centers \cite{wang2024testing} has been shown to substantially improve model robustness and generalization across demographic variations.

\textbf{Generative Data Augmentation} through generative adversarial networks (GANs) offers an alternative pathway for expanding limited training data. Transfer-GAN \cite{abbasi2023transfer} has been applied to augment the MHSMA \cite{javadi2019novel} and HuSHeM \cite{shaker2017dictionary} datasets, generating synthetically diverse but semantically consistent sperm images that improve classifier robustness to morphological variations. These generative approaches are particularly valuable for rare abnormality classes where natural sample availability is limited.

\subsection{Explainable AI and Clinical Integration}
\label{subsec:xai_clinical}

As deep learning models transition from research environments to clinical workflows, the demand for interpretability and transparent decision-making has become increasingly urgent. Explainable AI (XAI) techniques provide clinicians with understandable rationales for model predictions, fostering trust and enabling informed clinical decisions \cite{ma2023implementation,liang2023clinical}.

Gradient-based attribution methods such as Grad-CAM \cite{selvaraju2017gradcam} and its variants have been widely adopted to visualize the image regions influencing classification decisions, allowing clinicians to verify that models attend to clinically relevant morphological features rather than imaging artifacts. Integrated Gradients \cite{sundararajan2017axiomatic} provide axiomatically grounded attribution scores, while LIME \cite{ribeiro2016lime} yields locally faithful explanations through perturbation-based analysis; these techniques have subsequently been applied and adapted within medical imaging pipelines \cite{hao2023deep,hao2023towards}.

Beyond post-hoc explanation, clinical integration demands rigorous evaluation of model reliability and failure modes. AIOM \cite{chang2024p} presents a comprehensive platform for AI-assisted microscopy that incorporates quality assurance mechanisms and uncertainty quantification, enabling laboratory technicians to identify cases requiring expert review. Low-cost AI-integrated microscope systems \cite{phiphattanaphiphop2025low} have further democratized access to automated sperm analysis in resource-limited settings, though the validation of such systems under diverse operating conditions remains an active area of investigation.

Recent research has highlighted the risk of systematic overestimation in AI-assisted sperm quality assessment \cite{lafuente2024308}, where models may exhibit optimistic bias due to dataset shift or population stratification. Addressing these challenges requires the development of calibration-aware training procedures and continuous monitoring frameworks. The MIDRC-MetricTree \cite{drukker2024midrc} methodology provides a structured approach to evaluating AI performance across clinically relevant subgroups, ensuring equitable model behavior across diverse patient populations.

The convergence of explainable AI, federated learning, and multi-modal fusion points toward a future generation of sperm analysis systems that are not only accurate but also transparent, privacy-preserving, and clinically trustworthy. Realizing this vision will require sustained interdisciplinary collaboration between computer vision researchers, biomedical engineers, reproductive clinicians, and regulatory stakeholders to establish standardized validation protocols and deployment frameworks.

\section{Guidelines for Method Selection in Different Scenarios}
\label{sec:guidelines}

Selecting an appropriate computer vision or deep learning method for sperm analysis should not be regarded as a simple pursuit of the highest benchmark score. Instead, the choice of method should be determined by the clinical objective, imaging modality, annotation availability, computational resources, and deployment environment. In practical sperm analysis, different tasks impose distinct technical requirements. Sperm detection emphasizes localization accuracy and inference speed; motility analysis requires temporally consistent tracking; morphology assessment depends on reliable segmentation and fine-grained classification; DNA integrity and functional evaluation require biologically meaningful labels; and fertility prediction often demands the integration of heterogeneous visual, clinical, and molecular information. Therefore, a scenario-oriented method selection strategy is essential for translating computer vision and deep learning algorithms into robust, reproducible, and clinically useful sperm analysis systems.

For routine semen analysis and real-time computer-assisted sperm analysis (CASA), where the primary goal is to estimate sperm concentration and motility from microscopic videos, lightweight object detection models are generally preferred. YOLO-based detectors provide a favorable balance between inference speed and detection accuracy, making them suitable for high-throughput and real-time clinical workflows \cite{yuzkat2023detection,zhu2023yolov5s}. When sperm are sparsely distributed and the imaging condition is stable, single-stage detectors such as YOLOv5 or its lightweight variants can achieve efficient sperm localization with relatively low computational cost. However, in videos with dense sperm populations, low contrast, severe occlusion, or visual similarity between sperm and impurities, specialized tiny-object detectors or attention-enhanced YOLO variants are more appropriate because they are explicitly designed to handle small moving targets in complex microscopic backgrounds \cite{zou2022tod,zhu2023yolov5s}. In particularly challenging videos, ensemble or fusion-based detection strategies may further improve robustness by integrating complementary predictions from different detectors \cite{yuzkat2023detection}.

For motility analysis, detection alone is insufficient because clinically relevant kinetic parameters, including curvilinear velocity, straight-line velocity, average path velocity, and trajectory persistence, require temporally consistent sperm identities. In this scenario, detection-by-tracking frameworks are recommended. Pipelines that combine object detectors with tracking algorithms, such as YOLO-based detection followed by DeepSORT, are suitable when frame-level detections are reliable and individual sperm trajectories need to be reconstructed \cite{lopuran2025hybrid}. More advanced sperm-specific tracking systems, such as YOLOv8-based detectors integrated with dedicated tracking modules, are preferable for crowded microscopic videos in which identity switches and trajectory fragmentation are common \cite{zhang2024sperm}. If the application focuses on clinical motility grading rather than explicit trajectory reconstruction, end-to-end deep learning models can be used to directly predict WHO motility categories from video data \cite{haugen2023sperm,hicks2019machine}. Nevertheless, tracking-based methods remain more interpretable for clinical use because they provide explicit motion trajectories and quantitative kinetic parameters.

For sperm morphology segmentation, method selection should depend on whether the target is semantic segmentation, instance segmentation, or fine-grained anatomical part segmentation. U-Net and its variants remain strong baselines for biomedical image segmentation because they are computationally efficient and can perform well with limited training data \cite{melendez2021sperm,lv2022improved}. These models are particularly suitable for segmenting sperm heads or tails in relatively clean microscopic images. When sperm cells overlap or instance-level separation is required, Mask R-CNN or feature pyramid network-based methods are more appropriate because they jointly provide object localization and pixel-level masks \cite{marin2021impact,lewandowska2023ensembling}. For fine-grained segmentation of multiple sperm substructures, including the head, acrosome, nucleus, midpiece, and tail, task-specific or multi-stage frameworks are often required because these components differ substantially in size, contrast, and morphology \cite{movahed2018learning,movahed2019automatic}. Recent foundation-model-based approaches, such as Cascade SAM, are promising for reducing annotation burden, but their use in reproductive microscopy should be accompanied by domain adaptation, fine-tuning, and rigorous validation \cite{shi2024cs3}.

For sperm morphology classification, convolutional neural networks remain effective when datasets are moderate in size and diagnostic categories are clearly defined. ResNet-based, VGG-based, and attention-enhanced CNN models are suitable for distinguishing normal and abnormal sperm or identifying specific morphological defects \cite{kilicc2025deep,maalej2025advancements}. When the task involves subtle structural differences or multi-label defect recognition, attention mechanisms can improve performance by emphasizing informative regions such as the sperm head, neck, and tail \cite{kilicc2025deep}. Vision Transformers may be advantageous when sufficient training data or effective transfer learning is available, because their self-attention mechanism can capture global structural relationships that may be difficult for conventional CNNs to model \cite{aktas2023performance,aktas2025unveiling}. However, transformer-based models are generally more data- and computation-demanding; therefore, they should be selected cautiously for small datasets unless adequate pretraining, augmentation, and external validation are implemented.

For functional sperm assessment, including zona pellucida-binding ability, acrosome reaction status, stress response, and DNA integrity, the choice of method should be guided by the biological endpoint rather than image appearance alone. When single-sperm functional labels are available, supervised deep learning models can be trained to associate subtle morphological or optical features with fertilization-related outcomes \cite{leung2024p,leung2025automatic,park2023deep}. For sperm DNA fragmentation index prediction, deep learning models can automate labor-intensive halo or fluorescence-based assessment, but their reliability depends heavily on standardized imaging protocols and high-quality annotations \cite{kumar2023deep,mccallum2019deep}. For label-free functional evaluation, quantitative phase imaging combined with deep neural networks is attractive because it enables non-invasive assessment of sperm health and stress states \cite{butola2020high,butola2024quantitative}. In these scenarios, interpretability and biological plausibility are especially important, because algorithmic outputs may directly influence sperm selection in assisted reproductive technologies.

For multimodal fertility prediction, information fusion methods should be prioritized when image-based features alone are insufficient to explain clinical outcomes. Sperm quality is influenced by motility, morphology, DNA integrity, patient metadata, hormone levels, lifestyle factors, and laboratory conditions. Multimodal datasets such as VISEM demonstrate the value of combining semen videos with clinical and biological variables for reproductive health research \cite{haugen2019visem}. When heterogeneous data sources are available, early fusion, late fusion, or hybrid fusion strategies can be selected according to data alignment, modality completeness, and clinical interpretability. Early fusion is suitable when different modalities are synchronized and complete, whereas late fusion is more robust when modalities are partially missing or collected under different protocols. Hybrid fusion can further combine modality-specific feature extraction with global decision integration. From the perspective of information fusion, this direction is particularly important because it shifts sperm analysis from single-task visual recognition toward integrated decision-support systems that combine image, video, clinical, and molecular information.

For low-resource or point-of-care scenarios, computational efficiency, robustness, and hardware compatibility should be prioritized over architectural complexity. Lightweight CNNs, compact YOLO variants, MobileNet-based classifiers, pruning, quantization, and edge-device deployment are suitable strategies when the system must operate on portable microscopes, embedded processors, or low-cost clinical devices \cite{ilhan2020fully,phiphattanaphiphop2025low}. Such models may not always achieve the highest benchmark performance, but they can provide practical advantages in rural hospitals, small laboratories, and resource-limited fertility centers. In these settings, the most appropriate model is often the one that achieves stable performance under variable illumination, magnification, staining, and sample-preparation conditions while maintaining real-time inference speed.

For small datasets or scarce annotations, transfer learning, task-specific augmentation, semi-supervised learning, and foundation models should be considered. Many sperm image datasets remain limited in scale, highly task-specific, and affected by inter-expert variability. Transfer learning from large natural-image or biomedical datasets can improve convergence and generalization when only a small number of labeled sperm images are available \cite{marin2021impact}. Task-specific augmentation, including class-balanced sampling and morphology-preserving transformations, is useful for handling class imbalance and rare abnormalities \cite{jankowski2024learning}. Foundation models such as SAM may reduce the annotation burden, but their zero-shot performance should not be assumed to be clinically reliable without validation on sperm-specific images \cite{shi2024cs3}. When expert annotation is expensive, semi-supervised learning and active learning may provide efficient alternatives by prioritizing the most informative samples for manual labeling.

For cross-center generalization and clinical translation, robustness and validation criteria should be incorporated from the beginning of method selection. Models trained on a single microscope, staining protocol, laboratory environment, or patient population may fail when applied to external centers. Therefore, cross-dataset validation, external testing, calibration analysis, and subgroup performance evaluation are necessary before clinical adoption. Detection models should be assessed using IoU, AP, and mAP; segmentation models using Dice coefficient, IoU, and mIoU; tracking models using MOTA, MOTP, IDF1, identity switches, and fragmentation; and classification models using accuracy, precision, recall, F1-score, and ROC-AUC. Importantly, these metrics should be interpreted together with clinical relevance. For example, a detector with high mAP may still be inadequate for motility analysis if it produces unstable identities across frames, whereas a classifier with high accuracy may be clinically unsafe if it performs poorly on rare but important abnormal categories.

In summary, no single algorithm is optimal for all sperm analysis scenarios. YOLO-based detectors are appropriate for real-time detection and concentration estimation; detector-tracker pipelines are preferred for motility analysis; U-Net variants are suitable for efficient semantic segmentation; Mask R-CNN and related instance segmentation models are useful for overlapping sperm and part-level analysis; CNNs and attention-enhanced networks remain strong choices for morphology classification; transformer and foundation models are promising for complex or data-rich scenarios; and multimodal fusion models are essential when the goal is holistic fertility prediction. A rational method-selection framework should therefore balance task requirements, data characteristics, computational resources, interpretability, and clinical deployment constraints. Such a scenario-driven strategy can improve reproducibility, reduce inappropriate model use, and accelerate the integration of computer vision, deep learning, and information fusion into reliable sperm analysis systems.

\begin{figure}
    \centering
    \includegraphics[width=0.49\linewidth]{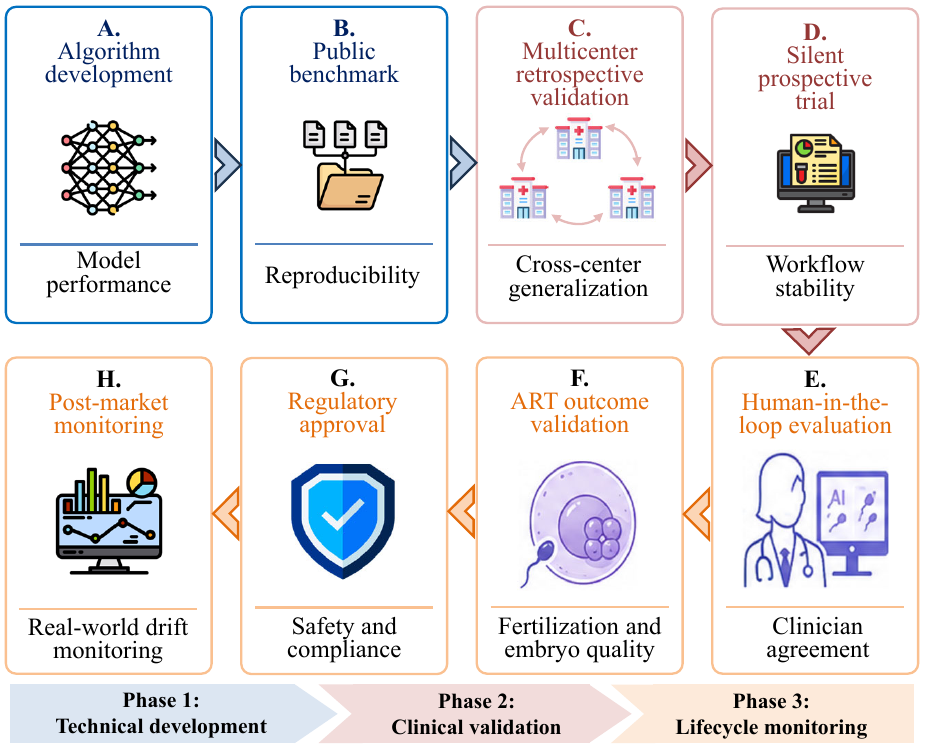}
    \vspace{-3mm}
    \caption{Clinical translation roadmap for AI-driven sperm analysis.}
    \label{fig:roadmap}
\end{figure}

\section{Clinical Transformation Challenges and Future Paths}
\label{sec:clinical}

Although computer vision and deep learning have substantially improved the objectivity, efficiency, and reproducibility of sperm analysis, their clinical transformation remains far from straightforward. In reproductive medicine, an AI model is not merely expected to detect sperm cells or classify morphology in isolated images; it must support a complex diagnostic and therapeutic pathway involving semen analysis, male infertility diagnosis, sperm selection, assisted reproductive technology (ART) planning, patient counseling, laboratory quality control, and long-term outcome monitoring. Therefore, the future of AI-driven sperm analysis should move beyond algorithm-centric performance optimization toward clinically integrated, multimodal, interpretable, and regulated decision-support systems.

A primary obstacle is the persistent gap between retrospective benchmark performance and real-world clinical utility. Many existing models are trained and evaluated on limited datasets collected from a single laboratory, microscope, staining protocol, camera system, or patient population. Such models may achieve high accuracy, Dice score, mAP, or MOTA under controlled experimental conditions, but their performance can degrade when exposed to different imaging devices, semen-preparation protocols, ethnic backgrounds, abstinence durations, sample viscosities, or disease etiologies. This issue is particularly important in sperm analysis because semen parameters are affected not only by imaging quality but also by biological variability, sample handling, operator procedures, and laboratory environment. Therefore, external validation across multiple centers, prospective testing, calibration analysis, and subgroup evaluation should become mandatory steps before deployment. Reporting frameworks such as TRIPOD+AI and DECIDE-AI provide useful methodological references for transparent reporting, early clinical evaluation, human--AI interaction assessment, and reproducibility of AI-based decision-support systems \cite{Collins2024TRIPODAI,Vasey2022DECIDEAI}.

Another major challenge lies in the definition of clinically meaningful endpoints. Current sperm AI models often focus on intermediate visual tasks, including sperm detection, segmentation, tracking, and morphology classification. However, reproductive clinicians are ultimately concerned with clinically actionable outcomes, such as fertilization probability, embryo development, implantation potential, miscarriage risk, live-birth rate, and the choice between natural conception, intrauterine insemination, IVF, or ICSI. A model that accurately detects sperm heads may not necessarily improve ART outcomes unless its outputs are linked to clinically relevant endpoints. The endpoint that matters can also lie upstream of semen analysis entirely: for men with non-obstructive azoospermia, deep learning has been explored to help identify sperm-bearing seminiferous tubules during microdissection testicular sperm extraction (M-TESE) \cite{elenkov2024260}, an intra-operative decision whose success directly conditions whether ICSI is possible at all. Such examples reinforce that clinical value is defined by the downstream reproductive decision, not by image-level accuracy in isolation. Future studies should therefore establish a hierarchical evidence chain from image-level performance to sample-level semen quality assessment, patient-level infertility diagnosis, and treatment-level reproductive outcomes. This requires longitudinal datasets that connect microscopic sperm images and videos with DNA fragmentation index, hormonal profiles, clinical history, ART procedures, embryo quality, pregnancy outcomes, and live-birth data \cite{haugen2019visem,mccallum2019deep}.

Multimodal data fusion represents one of the most important future paths for clinically meaningful sperm analysis. Male infertility is rarely determined by a single visual feature; instead, it reflects the interaction of sperm concentration, motility, morphology, DNA integrity, oxidative stress, endocrine status, genetic background, lifestyle factors, varicocele, infection, environmental exposure, and female-partner factors. Accordingly, future AI systems should integrate heterogeneous data streams, including microscopic images, time-lapse sperm videos, CASA-derived kinetic parameters, quantitative phase imaging, sperm DNA fragmentation assays, proteomic or genomic markers, electronic health records, and ART outcome data. Early fusion can be useful when modalities are synchronized at the single-sperm or sample level, whereas late fusion may be more robust when different modalities are collected at different time points or contain missing values. Hybrid fusion architectures, graph neural networks, and transformer-based cross-modal attention mechanisms may further enable structured reasoning over relationships among sperm-level, sample-level, and patient-level variables. From the perspective of \textit{Information Fusion}, this transition from single-modality visual recognition to multimodal reproductive decision support is a key direction for the field.

Recent progress in large language models (LLMs) and multimodal large language models (MLLMs) provides new opportunities for transforming sperm analysis from a stand-alone laboratory tool into an interactive clinical intelligence system. Medical LLMs such as Med-PaLM and Med-PaLM 2 have demonstrated the potential of large-scale language models for medical question answering and clinical reasoning \cite{Singhal2023MedPaLM,Singhal2023MedPaLM2}. More importantly, generalist biomedical AI systems such as Med-PaLMM and BiomedGPT suggest that a unified model can potentially process medical text, images, and genomic information within a single framework \cite{Tu2023MedPaLMM,Zhang2023BiomedGPT}. In the context of male infertility, an MLLM-based system could integrate semen-analysis reports, sperm microscopic images, motility videos, clinical history, endocrine test results, genetic findings, and ART records to generate structured diagnostic summaries, identify missing examinations, explain abnormal semen parameters, and recommend evidence-based next diagnostic steps. However, such systems should be positioned as clinician-facing decision-support tools rather than autonomous diagnostic agents, because reproductive treatment decisions involve uncertainty, patient preference, ethical considerations, and couple-level clinical context.

A particularly promising direction is multi-agent multimodal consultation for infertility management. Instead of relying on a single monolithic model, a multi-agent system could assign different roles to specialized agents: an imaging agent for sperm detection, tracking, and morphology interpretation; a genetics agent for DNA fragmentation, chromosomal, or sequencing-related information; a clinical reasoning agent for integrating male and female infertility factors; a guideline agent for checking consistency with WHO semen-analysis standards and ART recommendations; a counseling agent for generating patient-friendly explanations; and a safety agent for detecting uncertainty, bias, contraindications, or hallucinated recommendations. These agents could communicate through a shared patient representation and produce a consensus report under physician supervision. Recent surveys on LLM-based multi-agent systems indicate that multi-agent collaboration can improve complex problem solving, planning, and critique-based reasoning \cite{Guo2024MultiAgentSurvey}. For reproductive medicine, this architecture is attractive because infertility diagnosis is inherently multidisciplinary, involving andrologists, embryologists, reproductive endocrinologists, genetic counselors, laboratory technicians, and patients. Nevertheless, multi-agent systems also introduce new risks, including error propagation, overconfident consensus, hidden conflicts among agents, and difficulties in accountability. Therefore, agent communication, evidence grounding, uncertainty estimation, and audit trails must be explicitly designed.

Multimodal question answering and report generation are also likely to reshape the clinical workflow of sperm analysis. In a future AI-assisted andrology laboratory, clinicians may be able to ask questions such as ``Which samples show discordance between progressive motility and DNA integrity?'', ``Does this sperm population contain a higher proportion of tapered-head abnormalities?'', or ``Which sperm-selection strategy is most appropriate given the motility video, DFI result, and prior IVF failure?'' An MLLM could retrieve relevant visual evidence, quantify sperm-level features, summarize patient history, and generate a structured explanation. Similar paradigms have been explored in broader medical AI, including medical multimodal LLMs for image-text reasoning and multimodal generative copilots for pathology \cite{Liu2023MedicalMLLM,Lu2024PathChat,Yan2023GPT4VMedical}. However, direct adoption in sperm analysis requires domain-specific adaptation because sperm microscopy differs substantially from radiology, pathology, or dermatology in object scale, motion dynamics, optical artifacts, density variation, and the need for real-time single-cell assessment.

Interpretability and uncertainty awareness are indispensable for clinical acceptance. In sperm morphology or motility assessment, clinicians and embryologists need to understand why a model identifies a sperm as abnormal, why a trajectory is classified as progressive or non-progressive, or why a sample is predicted to have poor fertilization potential. Simple heatmaps may be insufficient because sperm analysis often depends on fine anatomical structures, including acrosome, nucleus, midpiece, and tail, as well as temporal motion patterns. Future systems should combine visual explanations, trajectory-based explanations, counterfactual examples, uncertainty scores, and rule-based clinical summaries. For example, a morphology model should not only output ``abnormal head'' but also indicate whether the decision is driven by head elongation, vacuole-like appearance, acrosomal defect, or segmentation uncertainty. Similarly, a motility model should report whether low confidence results from overlapping sperm, poor focus, low frame rate, or inconsistent tracking. Clinically meaningful explainable AI should therefore be evaluated not only by technical faithfulness but also by whether it improves clinician trust, error detection, and decision quality \cite{Jin2022ClinicalXAI}.

Privacy-preserving and collaborative learning will be essential for overcoming data scarcity and improving generalization. High-quality sperm datasets are difficult to build because they require expert annotation, standardized imaging, ethical approval, and linkage to sensitive reproductive outcomes. In addition, infertility data are highly private because they involve reproductive history, genetic information, and couple-level medical records. Federated learning offers a practical solution by enabling multiple fertility centers to collaboratively train models without directly sharing raw patient data. This is particularly valuable for building robust sperm-analysis models across diverse populations and laboratory protocols. However, federated learning is not a complete solution; non-IID data distributions, inconsistent annotation standards, communication cost, model drift, and privacy leakage through gradients remain important challenges. Future federated sperm-analysis networks should incorporate harmonized data dictionaries, shared annotation protocols, secure aggregation, differential privacy, and prospective monitoring of site-specific performance.

Clinical workflow integration is another major determinant of successful translation. AI systems must fit naturally into the daily routines of andrology laboratories and reproductive clinics. A clinically useful system should not increase workload by requiring excessive manual correction, complicated data upload, or time-consuming parameter adjustment. Instead, it should automatically acquire microscope images or videos, perform quality control, flag unreliable frames, quantify sperm parameters, generate structured reports, and integrate results into laboratory information systems or electronic health records. For embryologists, the system should support sperm selection without disrupting time-sensitive IVF or ICSI procedures. For clinicians, it should provide concise and interpretable summaries rather than overwhelming raw outputs. For patients, it should translate technical semen parameters into understandable explanations while avoiding deterministic or stigmatizing language. Therefore, future development should incorporate human-centered design, usability testing, and workflow simulation from the early stage of model development.

Regulatory and quality-management challenges must also be addressed. AI-assisted sperm analysis systems intended for clinical decision support may fall under medical-device or high-risk AI regulations depending on jurisdiction and intended use. In the European Union in particular, a software-plus-optics semen analyzer is likely to engage the In Vitro Diagnostic Medical Devices Regulation (IVDR, Regulation (EU) 2017/746) \cite{eu2017ivdr} in addition to the horizontal AI Act, since it produces information intended to inform clinical decisions from a specimen; the two frameworks impose overlapping but distinct obligations on clinical evidence, performance evaluation, and post-market surveillance. The EU AI Act emphasizes risk management, data quality, transparency, human oversight, and post-market monitoring for high-risk AI systems used in medical contexts \cite{EuropeanCommission2024AIHealthcare,EuropeanUnion2024AIAct}. Similarly, the FDA's principles for predetermined change control plans highlight the need to manage model updates, retraining, and real-world performance monitoring for machine-learning-enabled medical devices \cite{FDA2024PCCP}. These requirements are highly relevant to sperm AI because model performance may change when new microscopes, staining protocols, patient populations, or laboratory procedures are introduced. Therefore, deployable systems should include version control, locked and adaptive model modes, predefined update protocols, post-deployment drift detection, failure-mode analysis, and audit-ready documentation.

Ethical and social considerations are particularly important in infertility care. AI-generated assessments of sperm quality can influence emotionally sensitive decisions, including whether a couple proceeds to IVF, ICSI, surgical sperm retrieval, donor sperm, genetic testing, or lifestyle intervention. False-positive or false-negative predictions may lead to unnecessary treatment, delayed care, financial burden, psychological distress, or inappropriate reproductive counseling. Bias is another major concern. If training datasets underrepresent certain age groups, ethnic backgrounds, infertility etiologies, or low-resource clinical settings, AI systems may produce systematically different performance across populations. Moreover, patient-facing LLMs must be carefully constrained to avoid hallucinated medical advice, overconfident prognosis, or recommendations that exceed approved clinical guidelines. Ethical deployment should therefore include informed consent, clinician oversight, bias auditing, uncertainty communication, patient-centered explanation, and clear boundaries between education, decision support, and medical diagnosis.

Future clinical transformation should proceed through a staged roadmap. The first stage is technical standardization, including public benchmark datasets, unified annotation criteria, standardized imaging protocols, and task-specific evaluation metrics. The second stage is multicenter retrospective validation, where models are tested across laboratories, devices, populations, and sample-preparation protocols. The third stage is prospective silent-mode evaluation, in which AI outputs are generated but not used for clinical decisions, allowing real-world performance and failure modes to be quantified. The fourth stage is human-in-the-loop clinical evaluation, where embryologists and clinicians use AI outputs under supervision and the impact on diagnostic consistency, workflow efficiency, and decision quality is measured. The final stage is outcome-based clinical validation, where the effect of AI-assisted sperm assessment or selection on fertilization, embryo quality, pregnancy, and live-birth outcomes is evaluated. This staged pathway can reduce translational risk and align algorithm development with clinical evidence generation.

In summary, the future of AI-driven sperm analysis should be clinically grounded, multimodal, interpretable, privacy-preserving, and regulation-aware. Computer vision models will continue to provide the core capability for sperm detection, tracking, segmentation, and morphology analysis, but the next generation of systems will likely be built around multimodal fusion, medical foundation models, MLLM-based report generation, and multi-agent clinical reasoning. The central challenge is not only to build more accurate models, but also to ensure that these models improve reproductive decision-making, integrate safely into ART workflows, protect patient privacy, reduce rather than amplify clinical bias, and generate evidence that is meaningful to clinicians and patients. By following this path, AI-enabled sperm analysis can evolve from a laboratory automation tool into a trustworthy component of precision reproductive medicine.

\section{Discussion of Model Robustness and Domain Adaptation Strategies}
\label{sec:robustness}

The preceding sections have surveyed the remarkable progress in sperm analysis driven by computer vision and deep learning, from morphometric segmentation to motility assessment and multimodal integration. This section is deliberately positioned as the forward-looking, robustness-centred complement to the rest of the review: where Section~\ref{sec:emerging} organizes advanced paradigms by their current evidence status and Section~\ref{sec:clinical} lays out the clinical-translation pathway, here we examine the \emph{mechanisms} of failure under distribution shift and the architectural, generative, and causal strategies proposed to counter them. Where a technology recurs across sections (for example, federated learning or multimodal large language models), it is treated at a distinct level of analysis rather than repeated, and every method not yet validated on sperm data is flagged as transferable or prospective. Yet a sobering reality persists: the overwhelming majority of published models are validated on in-house datasets, under controlled laboratory conditions, with little evidence of generalization to new clinical environments. This section confronts the robustness challenge directly, offering not merely a catalog of domain adaptation techniques but a fundamental rethinking of what ``domain'' means in the context of sperm analysis. We argue that the field must move beyond surface-level visual adaptation toward deep biological and technical robustness, embracing emerging architectures, generative paradigms, and causal reasoning frameworks that have largely yet to be applied to andrology.

\subsection{Limitations of Current Approaches and New Perspectives}

Existing efforts to address domain shift in sperm analysis have followed predictable patterns. Transfer learning via ImageNet pre-training has served as the de facto starting point for virtually all convolutional approaches \cite{marin2021impact}. Federated domain adaptation methods such as FedDAG \cite{che2025feddag} have attempted to align feature distributions across centers without centralizing sensitive data. Cascade SAM with style transfer (CS3) \cite{shi2024cs3} and FACMIC \cite{wu2024facmic} have leveraged foundation models and frequency-domain adaptation to bridge visual gaps between microscopy systems. While these contributions represent genuine advances, they share a critical limitation: they conceptualize ``domain'' almost exclusively in terms of \textit{visual appearance}: staining protocols, illumination conditions, camera sensors, and magnification differences.

This visual-centric framing is fundamentally incomplete for sperm analysis. We propose that the field adopt a \textbf{multi-scale domain ontology} that recognizes domain shift operating at five interconnected levels: (1) the \textit{cellular level}, where individual sperm morphology varies with hormonal cycles, oxidative stress, and cryopreservation status \cite{butola2020high,kumar2023deep}; (2) the \textit{sample level}, where semen collection time, abstinence duration, and processing methods introduce systematic variation \cite{hsu2023live,you2021machine}; (3) the \textit{population level}, where ethnicity, age, body mass index, and environmental toxin exposure create distributional differences \cite{lafuente2024308}; (4) the \textit{temporal level}, where sperm quality within the same patient fluctuates across days, weeks, and seasons; and (5) the \textit{technical level}, encompassing microscope manufacturer, camera sensor, digital staining algorithms, and operator skill \cite{akal2023evaluation,wang2024testing}.

The critical insight is that current adaptation strategies conflate these distinct domain types. Visual adaptation techniques (style transfer, color normalization) may address technical-level shift but are powerless against biological covariates. Conversely, biological normalization (controlling for abstinence duration, cryopreservation method) does nothing to address camera sensor differences. A model trained on fertile patients in Barcelona and deployed in Shanghai faces not merely different microscopes, but different genetic backgrounds, dietary patterns, environmental exposures, and clinical protocols. Treating all of this as a single ``domain gap'' to be bridged by adversarial alignment is conceptually impoverished. We argue that \textbf{hierarchical adaptation strategies}, combining technical normalization at the input layer, biological conditioning at the feature level, and population-aware calibration at the output layer, represent a more principled path forward than any single adaptation mechanism.

\subsection{Deep Mechanisms of Domain Shift: Beyond Surface Variations}

To design effective robustness strategies, one must first understand the generative processes underlying domain shift. In sperm analysis, domain shift is not merely a statistical nuisance; it reflects genuine biological and technical variation with distinct causal structures.

At the \textbf{cellular level}, sperm morphology is exquisitely sensitive to physiological state. Oxidative stress alters membrane topology and tail curvature in ways that are visually detectable but biologically distinct from the morphological signatures of genetic abnormalities \cite{kumar2023deep}. Cryopreservation induces ice crystal damage that creates acrosomal irregularities easily mistaken for pathological forms by na"ive classifiers \cite{butola2020high}. A model trained on fresh semen that encounters cryopreserved samples for the first time may ``correctly'' flag these artifacts as abnormal morphology, not because it has learned the biological concept of cryodamage, but because it has learned a spurious correlation between certain texture patterns and the ``abnormal'' label.

At the \textbf{sample level}, the WHO laboratory manual specifies strict pre-analytical conditions, yet real-world compliance varies dramatically \cite{hsu2023live,you2021machine}. Abstinence duration affects sperm concentration and motility in predictable dose-response relationships. Collection time matters: circadian rhythms influence semen parameters, and samples collected in the evening may systematically differ from morning collections. Liquefaction time, processing temperature, and pH adjustments all introduce variation that a purely image-based model cannot explain without auxiliary clinical metadata.

The \textbf{population level} introduces perhaps the most challenging form of domain shift. Semen quality parameters differ significantly across ethnic groups, and reference ranges established in Caucasian populations may misclassify normozoospermic men from other ancestry groups \cite{lafuente2024308}. Environmental factors, endocrine-disrupting chemicals, air pollution, scrotal temperature from occupational exposures, create population-level trends that manifest as subtle distributional shifts in morphology and motility. A model trained in a low-pollution Nordic city and deployed in a high-pollution Asian metropolis may systematically under-diagnose oligospermia because its notion of ``normal concentration'' has been calibrated to a different environmental baseline.

The \textbf{temporal level} remains almost entirely unstudied in sperm AI. Unlike radiological images, which are typically acquired at discrete clinical encounters, semen analysis is often performed serially during fertility treatment. A model deployed for six months may encounter gradual shifts in patient population (seasonal variation in referrals), microscope calibration drift, and even software updates to the imaging system. These gradual shifts are invisible to standard domain adaptation frameworks, which assume a clear source-target dichotomy.

At the \textbf{technical level}, variation in microscope optics (phase-contrast vs. bright-field vs. differential interference contrast), camera sensor characteristics (CMOS vs. CCD, bit depth, dynamic range), and digital staining algorithms creates the most visually obvious form of domain shift \cite{akal2023evaluation,wang2024testing}. Yet even here, current approaches oversimplify: the point spread function of a high-NA objective interacts with the physical dimensions of subcellular structures (the acrosome is approximately 0.5 $\mu$m thick) in ways that cannot be fully normalized by simple color transfer or histogram matching.

\subsection{From Domain Adaptation to Domain Generalization: A Paradigm Shift}

The field of computer vision has undergone a conceptual evolution that sperm analysis has yet to fully embrace. The classical paradigm of \textit{Domain Adaptation} (DA) assumes access to unlabeled target domain data during training, an assumption that often fails in clinical practice, where a new hospital may not have begun data collection. \textit{Domain Generalization} (DG) relaxes this requirement, seeking to train models that generalize to \textit{unseen} domains without any target data \cite{seo2019learning}. More recently, the \textit{Foundation Model Era} has introduced pre-trained representations learned from billions of diverse images, offering implicit robustness through sheer exposure to variation \cite{Tu2023MedPaLMM}.

We advocate for a paradigm shift in sperm analysis from DA to DG, supported by three emerging technical directions. First, \textbf{causality-driven domain generalization} offers a principled framework for removing spurious correlations. The CROCODILE framework \cite{carloni2024crocodile} employs causal reasoning to identify features that are \textit{causally} related to the target variable (e.g., genuine morphological abnormalities) versus those that merely \textit{correlate} with it due to confounding (e.g., staining intensity differences between centers). By disentangling causal from spurious features through contrastive learning and prior knowledge injection, CROCODILE improves out-of-distribution performance at the cost of modest in-domain accuracy, a trade-off that is clinically defensible when patient safety is at stake. Applied to sperm analysis, such causal reasoning could distinguish, for instance, genuine head vacuolization (causally linked to DNA fragmentation) from fixation artifacts (spuriously correlated with particular laboratory protocols).

Second, \textbf{meta-learning for few-shot adaptation} \cite{farshad2022metamedseg} reframes domain adaptation as a learning-to-learn problem. Rather than training a single model to generalize across all domains, meta-learning algorithms like MAML seek an initialization from which the model can adapt to a new domain with minimal gradient steps and scarce labeled examples. For sperm analysis, this is transformative: a model meta-trained across multiple clinics could, in principle, adapt to a new microscope or patient population with as few as 10-20 annotated examples, rather than the thousands required for full fine-tuning. MetaMedSeg \cite{farshad2022metamedseg} has demonstrated this principle for volumetric organ segmentation, and extension to 2D sperm morphology classification is technically straightforward.

Third, \textbf{normalization-based strategies} offer lightweight yet effective alternatives to full architectural adaptation. Domain-Specific Optimized Normalization (DSON) \cite{seo2019learning} learns per-domain affine parameters and mixture weights between batch and instance normalization, effectively removing domain-specific style while preserving semantic discriminability. Domain-Specific Batch Normalization (DSBN) \cite{chang2019domain} maintains separate BN statistics per domain, allowing a single shared backbone to serve multiple centers with minimal overhead. For multi-center sperm analysis networks, these approaches offer an attractive middle ground: simpler than adversarial adaptation, yet more expressive than simple fine-tuning.

The unifying insight across these approaches is that \textbf{the future lies not in adapting to known domains but in building models that inherently do not rely on domain-specific spurious features}. A sperm morphology classifier that needs to be retrained for every new microscope is not a clinical tool; it is a research prototype. The goal must be models that extract biologically invariant representations, features that would remain stable even if the microscope, staining protocol, and patient population were all simultaneously changed.

The choice of backbone architecture interacts with this goal, particularly for high-resolution microscopy where the quadratic token complexity of Vision Transformers can become a bottleneck. Beyond efficient-attention and convolutional designs, state space models (SSMs) such as Vision Mamba \cite{zhu2024vision} have recently been explored as a linear-complexity alternative, and adaptations to medical imaging include classification (MedMamba \cite{yue2024medmamba}), hybrid CNN--SSM segmentation (U-Mamba \cite{ma2024umamba}), local--global token organization (LoG-VMamba \cite{dang2024logvmamba}), xLSTM-augmented variants (VMAXL-UNet \cite{zhong2025vmaxl}), and reconstruction under sparse sampling (MambaMIR \cite{zheng2024mambamir}). These are worth monitoring for sperm analysis, especially for full-field or high-frame-rate settings where token counts are large. However, their advantages over strong convolutional and efficient-attention baselines on natural images remain contested, their one-dimensional scan order fits the intrinsic two-dimensional structure of microscopy imperfectly, and, critically, no SSM has yet been validated on sperm data. We therefore regard SSMs as a promising but unproven backbone option rather than an established solution, with hybrid CNN--SSM designs the most defensible entry point should the field adopt them. Complementary backbone-level priors are also worth monitoring: group-equivariant convolutions (e.g., SE(3)-equivariant designs \cite{kuipers2023regular}) build geometric invariances directly into the architecture, which is attractive for microscopy where cell orientation is arbitrary, while masked multi-view transformer pre-training (e.g., SwinMM \cite{wang2023swinmm}) offers a label-efficient route to strong representations. As with SSMs, none of these has yet been validated on sperm data, and they should be framed as transferable rather than established.

\subsection{Generative AI and Synthetic Data: The New Data Augmentation Frontier}

Data scarcity is the perennial bottleneck in medical AI, and sperm analysis is no exception. Rare morphological abnormalities, amorphous heads, pinheads, coiled tails, cytoplasmic droplets, may represent less than 1\% of sperm in a given sample, yet they are precisely the clinically significant findings that models must detect. Traditional augmentation (rotation, flipping, color jittering) merely perturbs existing samples and cannot create fundamentally new morphological variants. Generative Adversarial Networks (GANs), such as Transfer-GAN \cite{abbasi2023transfer}, have offered a partial solution but suffer from training instability, mode collapse, and limited diversity.

\textbf{Diffusion models} represent a qualitative leap forward. Denoising Diffusion Probabilistic Models (DDPMs) and their latent variants (LDMs) learn to reverse a gradual noising process, enabling the generation of high-fidelity samples with superior diversity and training stability compared to GANs \cite{nazir2025diffusion, cao2025diffusion}. For sperm analysis, this capability addresses the long-tail problem directly: diffusion models can generate training examples of rare abnormalities that clinicians encounter too infrequently to build large datasets. Recent work by Nazir et al. \cite{nazir2025diffusion} has demonstrated text-guided diffusion-based augmentation for medical image segmentation, achieving 8-10\% Dice improvements by synthesizing rare pathological features on normal background images. Cao et al. \cite{cao2025diffusion} further validated diffusion models for medical image generation under low-resource computing conditions, confirming their practical viability for clinical laboratories without extensive GPU infrastructure.

The most exciting frontier is \textbf{text-guided cross-modal synthesis}. Latent diffusion models conditioned on clinical descriptions can generate semantically controlled variations: ``sperm with amorphous head and excess residual cytoplasm'' or ``tapered-head spermatozoon with bent midpiece.'' This enables the generation of training data for morphological classes that are underrepresented in real datasets. Moreover, cross-modal translation, converting between staining protocols (Papanicolaou to Diff-Quik), imaging modalities (bright-field to phase-contrast), or even simulating the effects of cryopreservation damage, offers a powerful form of \textit{adversarial domain augmentation} that trains models to be invariant to such transformations.

A particularly innovative direction is the MAGIC framework \cite{wang2025magic}, which uses Multimodal Large Language Models (MLLMs) to provide expert checklist-based feedback during diffusion model training. Rather than relying solely on pixel-level losses, MAGIC employs an MLLM evaluator (e.g., GPT-4V) to assess generated images against clinically defined criteria (``Does this generated sperm image show a correctly proportioned acrosome?''), translating expert knowledge into fine-tuning signals for the diffusion model. For sperm analysis, where expert embryologists possess nuanced, decades-honed criteria for morphological assessment, this human-in-the-loop paradigm could generate synthetic training data of unprecedented clinical fidelity. The synthetic data could then augment real datasets for rare morphological variants, creating robust classifiers without the years of data collection that would otherwise be required.

\subsection{Continual Learning and Test-Time Adaptation: Dynamic Robustness}

The prevailing assumption in sperm analysis AI is that deployment is a one-time event: train, validate, freeze, deploy. Clinical reality is far more dynamic. Microscopes require periodic recalibration that subtly shifts image characteristics. Seasonal variations in patient referrals alter the underlying population distribution. New staining protocols are introduced as laboratories update their procedures. A model frozen at deployment time gradually becomes a model calibrated to a world that no longer exists.

\textbf{Test-time training} (TTT) \cite{pan2025ttt} offers a compelling solution: the model adapts at inference time using self-supervised objectives on the unlabeled target sample, without requiring labeled target data. TTT-VNet \cite{pan2025ttt} has demonstrated this principle for medical image segmentation, using a auxiliary self-supervised rotation prediction task to adapt batch normalization statistics during inference. For sperm analysis, TTT could enable a deployed model to adapt to a new microscope within hours rather than weeks, continuously adjusting its internal representations as imaging conditions evolve.

For longer timescales, \textbf{continual learning frameworks} like MedCoSS \cite{ye2023medcoss} enable sequential learning across modalities without catastrophic forgetting. MedCoSS assigns different modality data to different training stages: X-ray, CT, MRI, pathology, forming a multi-stage pre-training process with rehearsal-based knowledge retention. The underlying principle is directly applicable to sperm analysis: a model could sequentially adapt to data from Clinic A (bright-field, Papanicolaou stain), then Clinic B (phase-contrast, Diff-Quik), then Clinic C (interference contrast, unstained), retaining knowledge from each while accommodating new visual domains. The k-means rehearsal sampling and feature distillation strategies in MedCoSS prevent the catastrophic forgetting that plagues naive sequential fine-tuning.

However, a fundamental tension exists between continual adaptation and regulatory requirements. The FDA's Predetermined Change Control Plan (PCCP) framework \cite{FDA2024PCCP} provides a pathway for locked AI systems with pre-specified modification protocols, but truly continual learning, where the model evolves autonomously, challenges the very notion of a ``locked'' version required for regulatory approval. A model that changes its weights with every new sample cannot be validated in the traditional sense. This tension between \textbf{dynamic robustness and regulatory stability} is one of the defining challenges for the next generation of clinical sperm analysis systems.

\subsection{Federated Privacy Computing: New Frameworks for Multi-Center Collaboration}

Federated Learning (FL) has emerged as the dominant paradigm for multi-center collaboration without data centralization, with applications in sperm analysis including FedDAG \cite{che2025feddag}, FACMIC \cite{wu2024facmic}, and MH-pFLGB \cite{xie2024mh}. Yet basic FL, aggregating model updates from distributed clients, is insufficient for the privacy and operational demands of fertility clinics.

\textbf{Differential privacy} (DP) provides mathematical guarantees against membership inference attacks by injecting calibrated noise into model updates. In sperm analysis, where the mere fact of a patient's visit to a fertility clinic may be sensitive information, DP is not merely a technical nicety but an ethical imperative. Secure aggregation protocols, which encrypt model updates such that the server can compute the aggregate but not inspect individual updates, complement DP by protecting against honest-but-curious server behavior. The combination of DP + secure aggregation creates a \textit{trusted compute environment} where no single party, not the participating clinic, not the aggregating server, not a potential adversary, can extract individual patient information.

\textbf{Split learning} offers an alternative architecture particularly suited to high-resolution microscopy: the neural network is split at a carefully chosen layer, with the early layers (processing raw pixel data) remaining on the local device, and only intermediate representations (which are far less informative than raw images) transmitted to a central server. For sperm analysis, this means that the sensitive microscope images never leave the clinic's premises; only high-level feature representations are shared. This architecture is especially relevant for high-resolution imaging where the bandwidth requirements of transmitting full-resolution raw images would be prohibitive.

Data quality assessment in FL presents unique challenges. In a centralized setting, data quality can be verified through direct inspection. In FL, the server has no access to raw data and cannot detect label noise, data poisoning, or systematic annotation errors from participating centers. Novel frameworks for \textit{contribution-based aggregation}, where client updates are weighted by estimated data quality rather than simple sample count, offer a partial solution. \textit{Incentive mechanisms}, cryptographically verifiable rewards for high-quality data contribution, will be essential for building sustainable clinical FL networks.

Cross-border fertility data sharing introduces legal complexities that technical solutions alone cannot resolve. Different jurisdictions maintain different regulations on reproductive data: HIPAA in the United States, GDPR in the European Union, and China's Personal Information Protection Law (PIPL) each impose distinct requirements on data processing, consent, and cross-border transfer. A federated sperm analysis network spanning multiple countries must align its technical privacy guarantees with the most restrictive applicable legal framework, creating what might be termed \textit{regulatory harmonization by design}.

\subsection{Causal Inference and Explainable Robustness}

The ``Clever Hans'' effect in medical AI \cite{hooper2024mitigating} refers to models learning spurious correlations, text markers on slides, hospital-specific artifacts, or subtle brightness cues, rather than biologically meaningful features. In sperm analysis, the risks are particularly acute. A morphology classifier might learn that a particular laboratory's images have slightly higher contrast due to their specific microscope settings, and exploit this as a proxy for ``normal'' versus ``abnormal'' classification based on which laboratory contributed which labels. Such a model would fail catastrophically when deployed to a clinic with different equipment.

\textbf{Causal discovery} offers a principled framework for identifying true causal factors of sperm quality versus confounders. By constructing causal graphs from observational data and expert knowledge, one can explicitly model the relationships between imaging conditions, biological state, and diagnostic labels. For instance, acrosome integrity is \textit{causally} related to fertilization potential; staining intensity is merely a \textit{confounder} that happens to correlate with acrosome visibility. A causal model would learn to condition on the acrosome structure itself, not on the staining intensity that makes it visible.

\textbf{Counterfactual reasoning} extends this framework by asking ``What would this sperm look like under different conditions?'' Counterfactual explanations, ``This sperm was classified as abnormal because of its head vacuolization; if the vacuolization were removed (counterfactually), it would be classified as normal'', provide clinically meaningful explanations that are grounded in biological concepts rather than pixel attributions. For sperm analysis, counterfactual generation could answer questions like ``What would this sperm look like under phase-contrast microscopy?'' or ``What would this sperm look like if cryopreservation damage were reversed?'' Such capabilities are essential for building trust with clinicians who need to understand \textit{why} a model made a particular prediction.

\textbf{Uncertainty quantification} serves as a critical robustness indicator. Rather than producing point estimates, models should express calibrated uncertainty about their predictions, flagging out-of-distribution samples for human review; post-hoc attribution methods such as Grad-CAM \cite{selvaraju2017gradcam}, Integrated Gradients \cite{sundararajan2017axiomatic}, and LIME \cite{ribeiro2016lime}, together with their medical-imaging adaptations \cite{hao2023deep,hao2023towards}, complement rather than replace such calibrated confidence estimates. Monte Carlo dropout, deep ensembles, and evidential learning each offer different trade-offs between computational cost and calibration quality. For sperm analysis, uncertainty-aware systems could automatically escalate ambiguous cases to senior embryologists, creating a human-AI collaborative workflow that leverages the speed of automation for clear cases and the judgment of experts for borderline ones.

The overarching insight is that \textbf{a robust sperm analysis system should not only predict correctly but also explain why and express uncertainty when it encounters unfamiliar patterns}. This requires moving beyond purely discriminative models toward \textit{generative-causal models} that can reason about the data generation process, simulate interventions, and acknowledge the limits of their knowledge.

\subsection{Large Language Models and Sperm Analysis: From Perception to Cognition}

The preceding discussion has focused on \textit{perceptual} AI: detecting, segmenting, and classifying sperm. The next frontier is \textit{cognitive} AI: reasoning about sperm quality in the broader context of patient history, partner factors, treatment options, and evidence-based guidelines.

Multimodal Large Language Models (MLLMs) such as Med-PaLMM \cite{Tu2023MedPaLMM} and BiomedGPT \cite{Zhang2023BiomedGPT} have demonstrated the ability to integrate text, images, and structured data within a unified reasoning framework. For sperm analysis, this could enable systems that process a microscopy image and the patient's clinical history (age, BMI, smoking status, previous fertility treatments) to generate a holistic assessment rather than an isolated morphological label. Liu et al. \cite{Liu2023MedicalMLLM} and Lu et al. \cite{Lu2024PathChat} have surveyed the rapidly evolving landscape of medical multimodal LLMs and generative AI copilots, identifying both the transformative potential and the unique challenges of medical applications.

\textbf{Multi-agent systems} \cite{Guo2024MultiAgentSurvey} offer a compelling architecture for clinical consultation. Rather than relying on a single monolithic model, a multi-agent system could comprise specialized agents: an \textit{imaging agent} for sperm detection and classification, a \textit{genetics agent} for integrating DNA fragmentation and chromosomal analysis results, a \textit{clinical reasoning agent} for interpreting findings in the context of patient history, and a \textit{guideline agent} for grounding recommendations in evidence-based protocols. These agents would communicate through a shared protocol, with an orchestrator agent synthesizing their outputs into a coherent clinical report. Guo et al. \cite{Guo2024MultiAgentSurvey} provide a comprehensive survey of multi-agent systems based on large language models, identifying medical consultation as a key application domain.

\textbf{Retrieval-Augmented Generation} (RAG) addresses one of the most critical limitations of LLMs in medicine: hallucination. By grounding LLM outputs in peer-reviewed literature, retrieving relevant passages from PubMed, the WHO laboratory manual, and clinical guidelines, RAG ensures that generated recommendations are evidence-based rather than confabulated. For sperm analysis, a RAG system could link morphological findings to the relevant literature (``This teratozoospermia pattern is associated with $XYZ$ gene mutation according to [retrieved study]''), providing both transparency and clinical utility.

Yet a critical caution is essential. \textbf{LLMs can hallucinate}, and in reproductive medicine, hallucinated recommendations could have profound consequences for patient care. Current MLLMs should be regarded as \textit{clinician-facing decision support tools}, not autonomous diagnostic agents. The physician must remain the final arbiter of clinical decisions, with AI serving as an intelligent assistant that surfaces relevant information, suggests considerations, and flags uncertainties. Yan et al. \cite{Yan2023GPT4VMedical} provide an early exploration of multimodal GPT systems for medical applications, highlighting both the promise and the need for careful validation.

\subsection{A Roadmap for Next-Generation Robust AI Sperm Analysis}

\begin{figure*}
    \centering
    \includegraphics[width=0.98\linewidth]{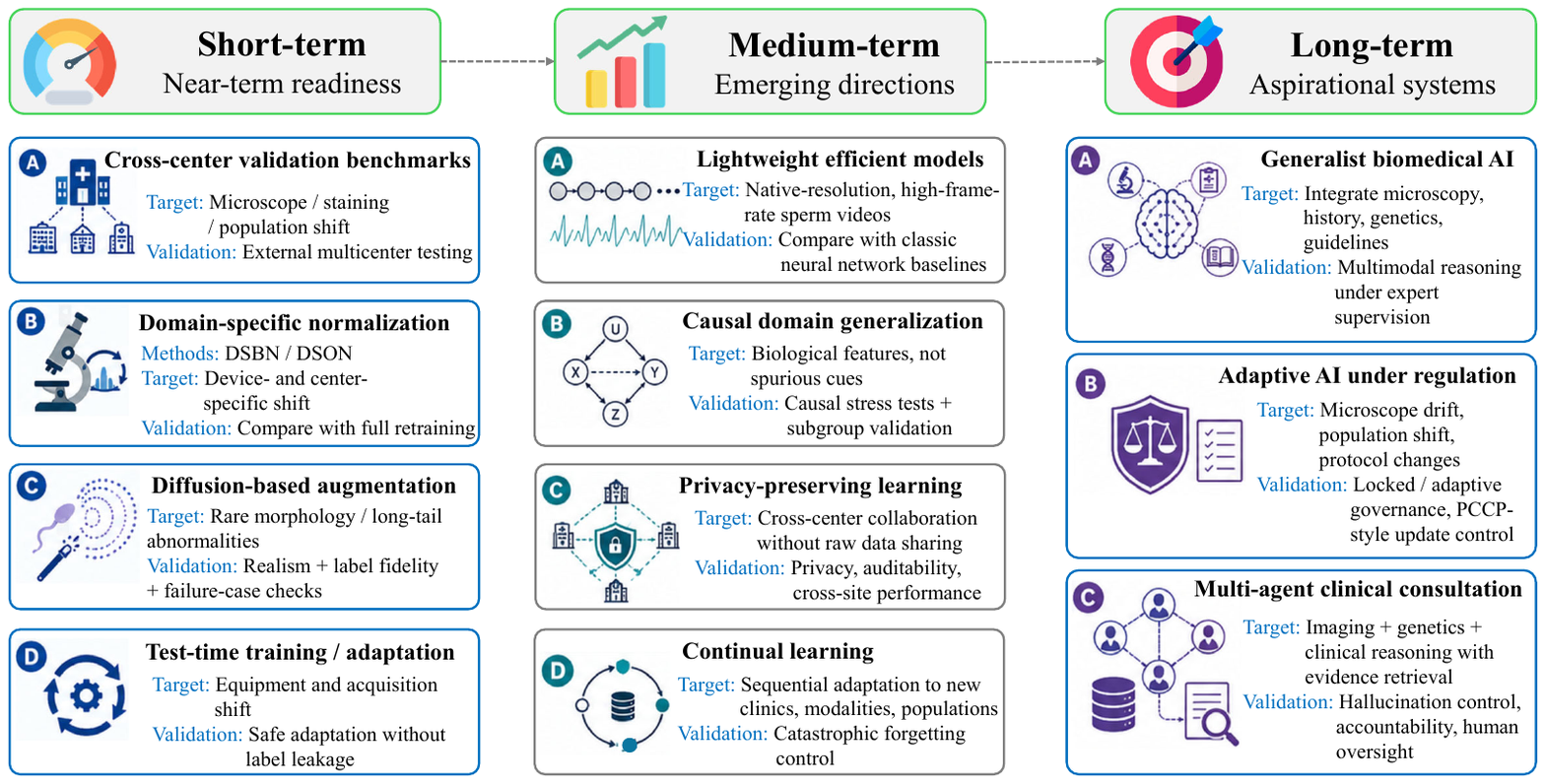}
    \vspace{-2.8cm}
    \caption{Roadmap for next-generation robust AI sperm analysis.}
    \label{fig:future_roadmap}
\end{figure*}

Synthesizing the foregoing discussion, as Figure \ref{fig:future_roadmap} presents, we outline an indicative, non-prescriptive roadmap for advancing robustness and domain adaptation in sperm-analysis AI. The horizons below reflect our reading of technical readiness rather than firm timelines, and consistent with the evidence tiers used throughout this review, most of the cited techniques are transferable from broader medical AI and still require dedicated validation on sperm data.

\textbf{Short-term (near-term readiness):} Diffusion-based data augmentation \cite{nazir2025diffusion, cao2025diffusion} could help address the long-tail problem of rare morphological abnormalities, and test-time training \cite{pan2025ttt} may offer a route to adaptation under equipment changes. A more immediate and lower-risk priority is the establishment of cross-center validation benchmarks that explicitly test for generalization across microscopes, staining protocols, and patient populations \cite{wang2024testing}, alongside lightweight domain-specific normalization strategies (DSON \cite{seo2019learning}, DSBN \cite{chang2019domain}) as alternatives to full model retraining.

\textbf{Medium-term (emerging directions):} SSM-based architectures \cite{ma2024umamba, dang2024logvmamba} are worth evaluating for native-resolution, high-frame-rate sperm analysis, where processing full microscope fields without tiling artifacts is attractive, though their advantage over strong convolutional and efficient-attention baselines remains to be demonstrated on sperm data. Causal domain-generalization frameworks \cite{carloni2024crocodile} may help models rely on biological rather than spurious features; federated learning networks \cite{che2025feddag,wu2024facmic,xie2024mh} with differential-privacy guarantees could enable regulatory-compliant collaboration across fertility centers; and continual learning \cite{ye2023medcoss} may support sequential adaptation to new clinics, modalities, and populations while mitigating catastrophic forgetting.

\textbf{Long-term (aspirational):} Generalist biomedical AI systems \cite{Tu2023MedPaLMM,Zhang2023BiomedGPT} with multimodal reasoning could, in principle, integrate microscopy, clinical history, genetic data, and evidence-based guidelines. Continual-learning systems that track gradual domain shifts (microscope drift, population changes, new clinical protocols) would need to operate within regulatory frameworks for adaptive AI \cite{FDA2024PCCP} and under appropriate human oversight rather than fully autonomously. Multi-agent clinical-consultation systems \cite{Guo2024MultiAgentSurvey} that synthesize imaging, genetic, and clinical reasoning into holistic assessments, ideally with retrieval-augmented grounding that traces recommendations back to peer-reviewed evidence, represent a plausible long-horizon goal, provided the well-documented risks of hallucination, overconfident consensus, and unclear accountability are resolved before any clinical use.

Taken together, the path from research prototype to clinical reality for sperm-analysis AI is unlikely to be traversed by incremental improvements to existing architectures alone. It calls for a broader reconceptualization of what ``domain'' means in reproductive medicine, drawing on causal reasoning, generative augmentation, state-space architectures, continual adaptation, and privacy-preserving collaboration. Several of the technologies surveyed here have the potential to be more than incremental, but whether they mature into dependable clinical tools, rather than remaining promising laboratory methods, will depend on rigorous, sperm-specific validation, transparent reporting, and sustained interdisciplinary collaboration between computer-vision researchers, reproductive clinicians, and regulatory stakeholders.

\section{Conclusion}
\label{sec:conclusion}

The field of AI-powered sperm analysis has undergone remarkable progress over the past decade, driven by rapid advancements in computer vision and deep learning. Traditional semen analysis, once limited by subjectivity and operator variability, is now being augmented by automated systems capable of delivering more objective and reproducible assessments, and emerging paradigms: multimodal data fusion, self-supervised learning, federated learning, and explainable AI, are beginning to address data scarcity, privacy protection, and clinical interpretability. Low-cost, integrated platforms further point toward democratized access to advanced reproductive diagnostics.

These advances should nonetheless be read with calibrated expectations, a theme we have maintained throughout this review. Reported accuracies are bounded by the reliability of the morphological and functional labels they are trained against, which is often only fair; headline percentages are drawn from heterogeneous datasets and protocols and are therefore not directly comparable across studies; a substantial share of the most striking clinical results currently exists only as conference abstracts or in animal models; and the kinematic parameters underpinning motility grading are only as trustworthy as the frame rate, calibration, and temperature control under which they were acquired. The principal open barriers are consequently not merely algorithmic but evidentiary: the absence of standardized, demographically documented, multi-centre datasets and a shared re-implementation benchmark; the difficulty of demonstrating genuine cross-population generalization; the need to anchor claims to clinically meaningful endpoints rather than image-level scores; and the integration of AI into existing clinical and regulatory workflows.

Moving forward, priority should therefore be given to multi-institutional data collaborations with harmonized annotation and disclosed population composition, to interpretable and uncertainty-aware frameworks whose outputs are constrained by clinical evidence, and to staged, outcome-based translational validation. Positioned this way, as trustworthy decision support under expert oversight rather than autonomous diagnosis, computer vision and deep learning can move sperm analysis from a laboratory automation tool toward a dependable component of precision reproductive medicine, advancing the management of male infertility while remaining honest about what current evidence does and does not establish.

\printcredits

\bibliographystyle{cas-model2-names}

\bibliography{cas-refs}





\end{document}